\documentclass[letterpaper, 10 pt, conference]{ieeeconf}  
\usepackage{amsmath}
\usepackage{amssymb} 
\usepackage{algorithm} 
\usepackage{algpseudocode}
\usepackage{multirow}
\usepackage{booktabs,subcaption}
\usepackage{graphicx}
\usepackage{wrapfig}
\usepackage{array}
\usepackage{hyperref}
\newcommand{\pmsd}[1]{\,{\scriptscriptstyle\pm #1}}

\IEEEoverridecommandlockouts                              

\title{\LARGE \bf
Test-Time Adaptation of Manipulation Policies \\Under Actuator Degradation
}

\author{Som Sagar$^{1}$, Ransalu Senanayake$^{1}$%
\thanks{$^{1}$School of Computing and Augmented Intelligence, Arizona State University, Tempe, Arizona
        {\tt\small  <ssagar6, ransalu>@asu.edu}}%
}

\begin{document}

\maketitle
\thispagestyle{empty}
\pagestyle{empty}

\begin{abstract}
Robot manipulation policies are usually trained under the assumption that a commanded action produces the same motion as it did during training even after hours of operation. Real hardware violates this assumption as the motors gradually heat up, current saturates near contact, voltage sags under load, thus the same policy action can produce a weaker, delayed, or noisier motion. These conditions are already measured by onboard telemetry, such as joint temperature, motor current, and supply voltage, yet this signal is typically used only for logging or safety checks rather than policy adaptation. We introduce \emph{Telemetry-Aware Action Rectification} (TeAR), a policy-agnostic method that turns a frozen manipulation policy into a telemetry-conditioned policy by rectifying its outgoing action before it reaches the low-level controller. TeAR learns a lightweight Transformer that combines the proposed action with live actuator telemetry and amplifies, damps, or biases individual action components. We evaluate TeAR across 18 policy--task pairs spanning 8 policy families and 5 manipulation tasks. In an additional paired evaluation with degradation-model mismatch, TeAR achieves 31.8\% success, compared with 25.6\% for the base policy and 30.6\% for an assumed-model inverse. On a physical arm, TeAR improves success under heating by 10--15\% without on-robot fine-tuning.

Code: \href{https://github.com/somsagar07/TeAR-Telemetry-Aware-Action-Rectification}{https://github.com/somsagar07/TeAR-Telemetry-Aware-Action-Rectification}.
\end{abstract}

\section{INTRODUCTION}

Pre-trained manipulation policies, from small behavior-cloned models to large vision-language-action (VLA) models, are often deployed without explicitly accounting for changes in the action-to-motion relationship. During operation, motor heating, current limits, and supply-voltage variation can change the actuator response that realizes a policy command. These effects need not produce immediate hardware faults, but they can alter the executed motion continuously while the robot remains operational. As a result, the policy may continue to output actions that would be appropriate for the actuator response seen during training, while the deployed hardware executes those same actions with reduced magnitude, added delay, or increased noise.

\begin{figure}[t]
  \centering
  \includegraphics[width=1\linewidth]{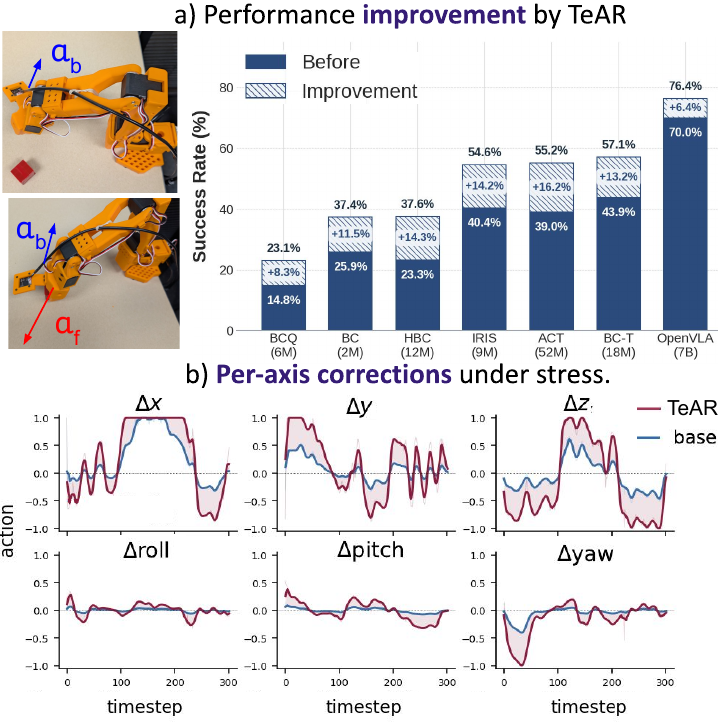}
  \vspace{-1.5em}
\caption{\textbf{(a)} Mean stressed success rate for 7 base policies ($2$M to $7$B parameters): TeAR (hatched) adds $+12\%$ on average over the frozen base (solid). \textbf{(b)} Over one stressed episode, TeAR (red) corrects the base intent (blue) across axes; the inset shows the rectification $a_b \!\to\! a_f$ on a SO-101 arm.}
  \label{fig:intro_1}
  \vspace{-1.4em}
\end{figure}

These effects are especially consequential in manipulation, where success depends on precise motion and coordinated contact forces~\cite{zhao2023act}, and execution delays can disrupt otherwise effective policies~\cite{black2026real}. Temperature-dependent changes in actuator torque response further challenge the assumption that a fixed command produces consistent motion~\cite{youn2024thermal}. Dynamics randomization exposes policies to varied physical parameters during training~\cite{peng2018simtoreal}, but does not by itself provide deployment-time measurements of actuator conditions. Without explicit telemetry inputs, a policy must infer these changes from their effects on task observations. TeAR instead uses measured actuator state to condition its action corrections. Figure~\ref{fig:intro_1} illustrates these corrections and the resulting gains under stress.

These deployment-time state values are already measured by most standard servo communication buses, which report joint temperature, motor current, and supply voltage at each control step. These measurements provide a compact description of the actuator conditions. Although such signals are routinely exposed by manipulator firmware, robot learning systems typically treat them as diagnostic logs or discrete safety thresholds rather than as continuous conditioning variables for the policy. We instead treat telemetry as a learning signal: a deployed action should depend not only on the scene and task, but also on the current actuator state under which the command will be executed.

\begin{figure*}[t]
  \centering
  \vspace{0.5em}
  \includegraphics[width=1\textwidth]{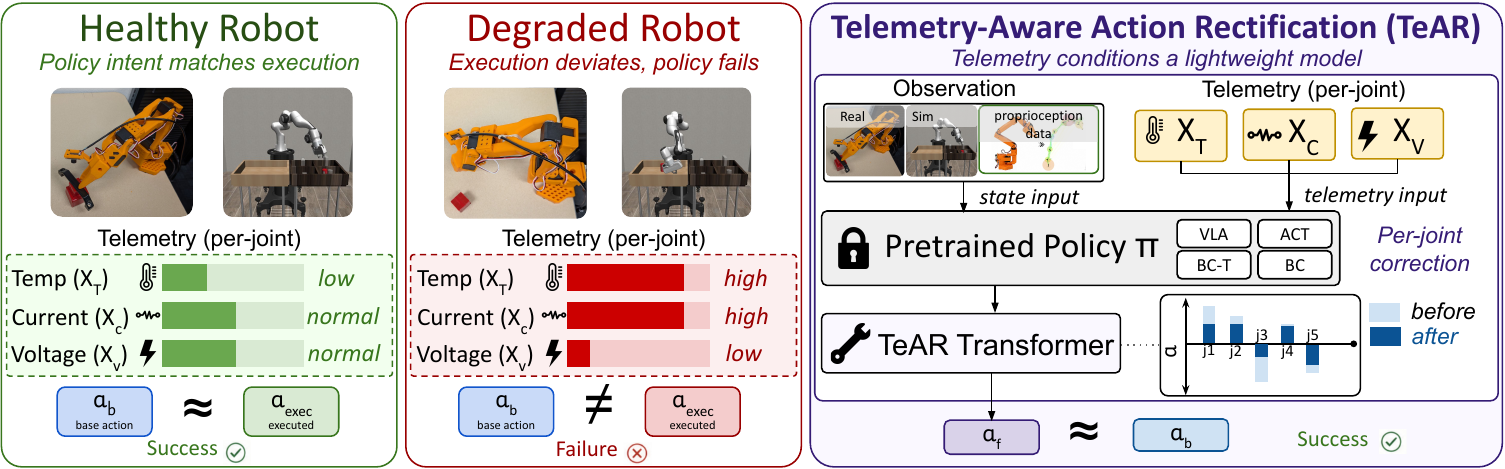}
\caption{\textbf{Telemetry-Aware Action Rectification.}  A policy assumes its command $a_b$ and executes: on healthy hardware it succeeds (left), but under joint heating, or voltage sag the executed action $a_{\text{exec}}$ drifts from $a_b$ and the task fails (middle). TeAR (right) reads per-joint temperature, current, and voltage and rewrites each command into $a_f$ so execution recovers.}
\vspace{-1.2em}
  \label{fig:intro}
\end{figure*}


We introduce \emph{Telemetry-Aware Action Rectification} (TeAR), a method for making a manipulation policy hardware-aware at test time (Fig.~\ref{fig:intro}). Our key design choice is to place the correction on the actuator response rather than on the policy internals since training with hardware state such as voltage or temperature is difficult, as it is not easily controllable like the environment state. Thus, TeAR reads the action proposed by the base policy together with live joint temperature, motor current, and supply voltage, and uses a Transformer to reshape the action before execution. Under actuator stress, this rectification can amplify, damp, or bias individual action components according to the sensed hardware state. Because the correction does not touch the base policy, the adapter can be trained in simulation against a physics-derived target, without environment roll-outs or base-policy fine-tuning, and transferred to real hardware.

The paper makes three contributions:
\begin{enumerate}
    \item \textbf{Telemetry-conditioned rectification.} We formulate actuator telemetry as test-time context for correcting frozen-policy actions at the policy-controller interface.
    \item \textbf{Efficient action adaptation.} We introduce a gated Transformer adapter trained from a physics-derived target, without base-policy fine-tuning.
    \item \textbf{Experimental validation.} We evaluate TeAR across 18 policy--task pairs, assess degradation-model mismatch, demonstrate thermal transfer to a physical arm.
\end{enumerate}

\section{Related Works}

\textbf{Adapters and residual wrappers for frozen policies.} Residual policy learning~\cite{silver2018residual,johannink2019residual} freezes a base controller and learns a bounded additive correction with RL, both to repair miscalibration and to recover sample efficiency on contact-rich tasks. Recent extensions adapt frozen imitation and diffusion policies through closed-loop residuals~\cite{ankile2024imitation,yuan2024policy} or by running RL in the latent-noise space of a frozen sampler~\cite{wagenmaker2025steering}. The vision-language-action literature has, in parallel, settled on parameter-efficient finetuning: OpenVLA~\cite{kim2024openvla} ships with LoRA, OpenVLA-OFT~\cite{kim2025finetuning} adds a continuous action head, $\pi_0$~\cite{black2024pi0} attaches a flow-matching expert, and Octo~\cite{octo2024} supplies a transformer-diffusion backbone designed for new heads. Across both families the correction is either entangled with the base architecture or conditioned on the same observations the base already consumes, and TRANSIC~\cite{jiang2024transic} relies on online human teleoperation as its corrective signal. TeAR provides a common telemetry-conditioned action interface for frozen policies. Its use across policy families requires compatible state/action conventions and an appropriate adapter checkpoint.

\textbf{Hardware telemetry and adaptive policy conditioning.}
Adapting robot behavior to changing dynamics often relies on inferring hidden system properties from interaction history. RMA~\cite{kumar2021rma} follows this approach by training a policy with privileged environment information and learning an adaptation module that estimates it from proprioceptive history. Related approaches use online system identification and meta-RL to adapt behavior from observed transitions~\cite{yu2017upops,nagabandi2019meta}. Explicit measurements of actuator operating conditions offer complementary context for adaptation~\cite{rugh2000research}. Temperature, current, and voltage provide such measurements, although they do not directly identify the resulting motion response. Relating these signals to motion still requires a model or learned prior, and ordinary task execution may provide limited coverage of operating conditions. TeAR learns telemetry-conditioned action corrections offline and applies them without updating the deployed policy.

\textbf{Robustness and degradation in manipulation.} Robustness to hardware variation has almost always been delivered at training time. Domain randomization~\cite{tobin2017domain,peng2018simtoreal} compresses an entire distribution of motor friction, joint damping, link inertias, and gear backlash into a single conservative policy, with adaptive and adversarial extensions that preserve the central premise~\cite{akkaya2019solving,pinto2017robust}. Fault-tolerant learning targets the discrete failure regime: ACDR~\cite{okamoto2021reinforcement} trains a PPO policy under a curriculum that randomly disables joints. Direct evidence that this strategy is insufficient at the VLA scale comes from the OpenVLA-OFT authors themselves, who attribute a sharp drop in bimanual success between training and a re-evaluation weeks later to wear in a small number of joints~\cite{kim2025finetuning}. TeAR trains an adapter conditioned on measured hardware telemetry, so the deployed behavior tracks the actual motor state rather than being fixed in advance.

\begin{figure*}[t]
\centering
\vspace{0.5em}
\includegraphics[width=0.95\linewidth]{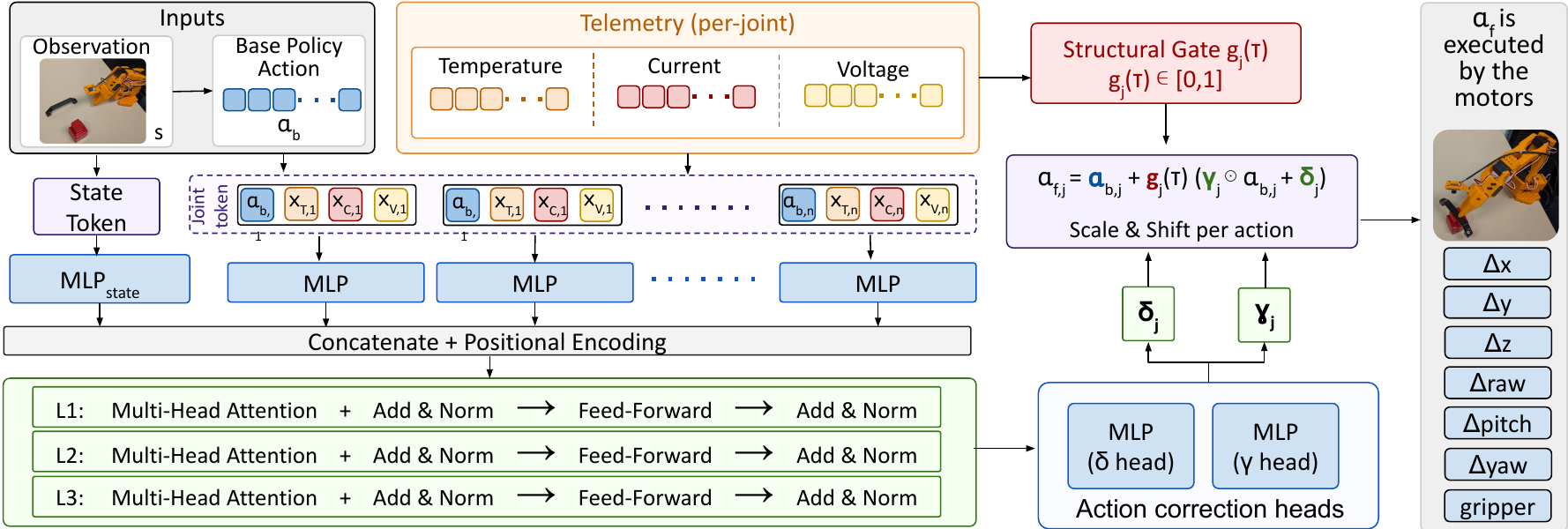}
\caption{A frozen policy proposes an action vector. Each action component is paired with indexed actuator telemetry to form a telemetry--action token. Together with a robot-state token, these inputs are processed by a three-layer Transformer. Two output heads predict multiplicative and additive corrections, activated by a telemetry-dependent gate.}
\label{fig:tam_arch}
\end{figure*}

\begin{figure}[t]
\centering
\includegraphics[width=1\linewidth]{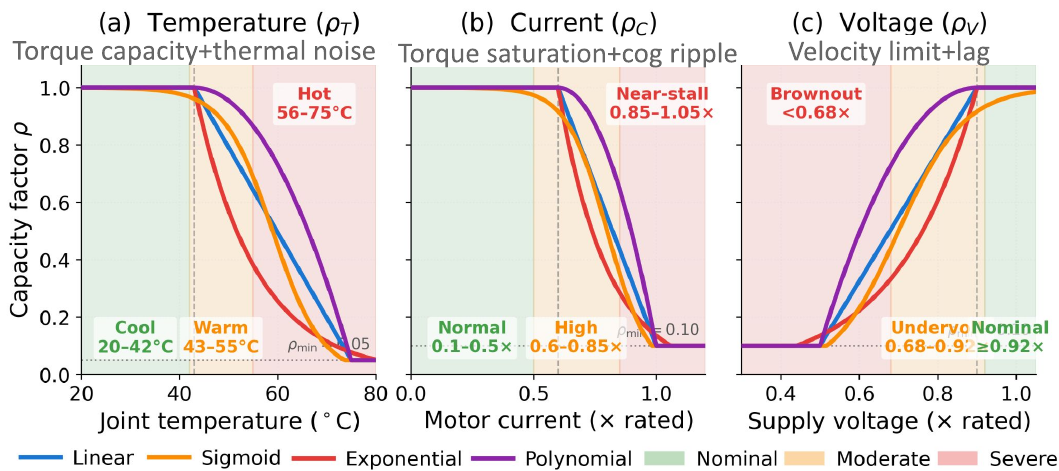}
\vspace{-1em}
\caption{\textbf{Degradation model.} Per-joint capacity factor $\rho \in [\rho_{\min}, 1]$ for each telemetry channel: (a) temperature reduces torque capacity, (b) current induces saturation and cogging ripple, and (c) voltage limits attainable velocity. Shaded bands mark the physical operating regimes.}
\vspace{-2em}
\label{fig:degcurves}
\end{figure}

\section{Multi-Channel Degradation Model}
\label{sec:degradation}
We formalize deployment-time actuator degradation as a telemetry-conditioned perturbation of the policy action, giving controlled and stress conditions in simulation. Given per-joint temperature, current, and voltage measurements, our degradation wrapper computes a capacity factor $\rho_c(x_{c,j})\in(0,1]$ for each channel $c\in\{T,C,V\}$ and joint $j$. A factor of one represents nominal capacity; smaller values represent reduced capacity under stress. The telemetry-dependent curves in Fig.~\ref{fig:degcurves} provide controlled simulation conditions rather than calibrated models of a particular robot.

In the nominal linear model, temperature capacity decreases from 1 above $43^\circ$C to a floor of 0.05 at $75^\circ$C; normalized current capacity decreases from 1 above 0.60 to 0.10 at 1.0; voltage capacity decreases from 1 below 0.90 to 0.10 at 0.50. Current and voltage are normalized model inputs, not amperes and volts.

Because the evaluated policies output operational-space control (OSC)~\cite{khatib1987osc}, we map joint-level capacities to actions. For each channel, we average joint capacities within two predefined, overlapping groups: $G_p$ for translation and $G_r$ for rotation. The combined capacity for each group is
\begin{equation}
q_k = \prod_{c\in\{T,C,V\}}
\left(\frac{1}{|G_k|}\sum_{j\in G_k}\rho_c(x_{c,j})\right),
\quad k\in\{p,r\}.
\label{eq:grouped}
\end{equation}
The three translation coordinates are scaled by $q_p$ and the three rotation coordinates by $q_r$; the gripper is unchanged. This grouping defines a repeatable action-level stress model rather than a physical mapping from joint torque limits to Cartesian motion.

Each channel also introduces its associated perturbation: temperature and voltage add noise, while current adds ripple. The wrapper applies the channels sequentially,
\begin{equation}
a_{\mathrm{exec}}=\mathcal{V}\!\left(\mathcal{C}\!\left(\mathcal{T}(a_f)\right)\right),
\label{eq:degcomposition}
\end{equation}
and passes the resulting action to the low-level controller. Ignoring these perturbations, the positive capacity factors make the scaling algebraically invertible, although action bounds can prevent exact compensation. Section~\ref{sec:training} describes the approximate inverse-based targets used to train TeAR.

\section{TeAR Framework}
\label{sec:method}
\emph{Telemetry-Aware Action Rectification} (TeAR) wraps a manipulation policy with a lightweight adapter that reads the three telemetry channels of Section~\ref{sec:degradation} and corrects its outgoing action. The adapter is inserted between the base policy and the low-level controller, producing an action of the same dimensionality as the base command without changing the policy internals.

\textbf{Setup.} A frozen base policy maps task observations $o$ to a normalized action $a_b=\pi_b(o)\in[-1,1]^{d_a}$. The adapter additionally receives a robot-state vector $s$ and actuator telemetry $\boldsymbol\tau=(\mathbf T,\mathbf C,\mathbf V)$. In the controlled simulation study, $d_a=7$: three translation coordinates, three rotation coordinates, and one gripper coordinate for OSC. Action coordinates are not joint torques. TeAR returns an action of the same dimension for the controller.

\subsection{Adapter Transformer Architecture}
\label{sec:arch}

We instantiate the adapter $\pi_\theta$ as a lightweight Transformer (Fig.~\ref{fig:tam_arch}). For each action index $j$, we form a four-dimensional \emph{telemetry-action token} $(a_{b,j},T_j,C_j,V_j)$ that pairs the proposed action component with the telemetry at the same index. A shared learned projection maps these tokens to $\mathbb{R}^{d_{\mathrm{model}}}$, and learned positional embeddings distinguish their indices. The robot-state vector $s$ is projected into an additional \emph{state token}, yielding a sequence of $d_a+1$ tokens. For the seven-dimensional OSC interface, action coordinate $j$ is paired with telemetry entry $j$ in the fixed input ordering. This pairing defines the representation rather than a physical correspondence between Cartesian axes and joints; self-attention makes the full telemetry vector available to every action token.

A three-layer Transformer encoder with $d_{\mathrm{model}}=128$ and four attention heads processes the sequence. Self-attention allows the correction at each action index to depend on the other telemetry-action tokens and the robot state. Two zero-initialized linear output heads map each action token to logits $u_j$ and $v_j$. A telemetry gate $g_j(\boldsymbol\tau)\in[0,1]$ then modulates the multiplicative gain and additive residual:
\begin{align}
\gamma_j&=1+g_j(\boldsymbol\tau)\gamma_{\max}\tanh u_j,\nonumber\\
\delta_j&=g_j(\boldsymbol\tau)\alpha\tanh v_j.
\end{align}
Here, $\gamma_{\max}$ bounds the gain's deviation from one, and $\alpha$ bounds the additive correction. The corrected action is
\begin{equation}
a_{f,j}=\mathrm{clip}(\gamma_j a_{b,j}+\delta_j,-1,1).
\label{eq:tamout}
\end{equation}
The gain scales the base command, while the residual provides an additive shift conditioned on the action, state, and telemetry. Together, they allow TeAR to amplify, damp, or bias individual action components. The final effect depends on both terms and action clipping; a gain below one need not reduce the output magnitude when the additive term reinforces the command.

We separate the decision to activate correction from the correction itself. The telemetry gate defines a nominal operating region in which the base action is preserved, while TeAR learns how to modify commands outside that region. For each channel $c$, we compute a normalized severity:
\begin{align}
z_{c,j}&=\mathrm{clip}\!\left(\frac{s_c(x_{c,j}-b_c)}{w_c},\,0,\,1\right),\nonumber\\
g_j(\boldsymbol\tau)&=\min\!\left\{1,\sum_{c\in\{T,C,V\}}h(z_{c,j})\right\}.
\label{eq:gate}
\end{align}
Here, $x_{T,j}=T_j$, $x_{C,j}=C_j$, and $x_{V,j}=V_j$; $b_c$ is the nominal-band boundary, $w_c>0$ is the transition width, and $s_T=s_C=1$, $s_V=-1$ specifies whether increasing or decreasing telemetry indicates stress. We use $h(z)=3z^2-2z^3$, which increases smoothly from zero to one with zero slope at both endpoints. Summing channel responses allows stress in any channel to activate correction, while clipping bounds the gate by one.

When all measured channels at index $j$ lie within their nominal bands, $g_j=0$. Consequently, $\gamma_j=1$, $\delta_j=0$, and Eq.~\ref{eq:tamout} gives $a_{f,j}=a_{b,j}$ exactly for finite network outputs and base actions in $[-1,1]$. Nominal pass-through is therefore enforced by the architecture rather than learned from data. This property concerns action identity under nominal measured telemetry; it does not establish closed-loop stability or safety under stress. The evaluated adapter applies this mechanism to all action components.

\subsection{Training: severity-augmented supervised fine-tuning}
\label{sec:training}

The supervised correction targets are constructed using an index-wise approximation to the degradation model in Section~\ref{sec:degradation}. At each action index $j$, the product $\rho_j(\boldsymbol{\tau})=\rho_T(x_{T,j})\rho_C(x_{C,j})\rho_V(x_{V,j})$ describes the capacity used for supervision. Unlike the grouped capacities in Eq.~\ref{eq:grouped}, this target uses telemetry at the same index and omits additive perturbations. Given a demonstration action $a_d$ and a sampled telemetry profile $\boldsymbol{\tau}$, we compensate for this approximate capacity loss and enforce the action bounds:
\begin{equation}
a^\star(a_d, \boldsymbol{\tau}) = \mathrm{clip}\!\Bigl(\,a_d \,\oslash\, \max\!\bigl(\boldsymbol{\rho}(\boldsymbol{\tau}),\, \rho_{\min}\bigr),\; -1,\; 1\,\Bigr),
\label{eq:sftarget}
\end{equation}
where $\oslash$ denotes element-wise division and the floor $\rho_{\min}=0.05$ limits amplification before clipping. We train $\pi_\theta$ to predict $a^\star$ from $(s,a_d,\boldsymbol{\tau})$ under an L1 loss, optionally adding noise to the demonstration-action input while keeping the target unchanged. At deployment, the action input is the frozen policy's proposed command $a_b=\pi_b(o)$. Both TeAR and analytic compensation rely on a prior relating telemetry to actuator response. TeAR learns a bounded correction conditioned on state, action, and telemetry, while the assumed-model inverse directly applies the nominal capacity curves. Neither receives the true deployment-time capacities. Section~\ref{sec:mismatch} tests whether the learned correction transfers when the actuator response differs from that prior.

\textbf{Severity curriculum.} To expose the adapter to the full telemetry range without on-policy rollouts, we stratify training samples into four tiers: clean, mild, moderate, and severe. For each nominal demonstration action, we sample one telemetry profile from each tier and compute its corresponding target $a^\star$, yielding four supervised examples per demonstration. On clean-tier samples no compensation is needed: $\boldsymbol{\rho}=\mathbf{1}$ and $a^\star=a_d$; these examples anchor the supervised dataset at the nominal regime, while the structural gate provides the pass-through guarantee at inference.

\begin{table*}[t]
\centering
\small
\renewcommand{\arraystretch}{1.08}
\vspace{0.2em}
\caption{Task success (\%) for 18 policy--task pairs across eight policy families and five tasks. Stressed-condition cells report base$\rightarrow$TeAR success; cool reports nominal success. Hot, stall, and brown apply temperature, current, and voltage stress individually; the moderate profiles activate the channels indicated by their subscripts. $\Delta$ reports the mean stressed-condition gain over the base, with the reported standard deviation across evaluation seeds.}
\label{tab:rq1_main}
\vspace{-0.5em}
\resizebox{\textwidth}{!}{%
\begin{tabular}{@{}>{\centering\arraybackslash}m{0.85cm}|l|*{8}{c}|r@{}}
\toprule
Task & Base policy & cool & hot & stall & brown & $\mathcal{T}_{\text{mod}}$ & $\mathcal{TC}_{\text{mod}}$ & $\mathcal{TV}_{\text{mod}}$ & $\mathcal{TCV}_{\text{mod}}$ & $\Delta(\uparrow) \pm \sigma $ \\
\midrule
\multirow{5}{*}{\rotatebox{90}{\textit{Lift}}} & BC ($\sim$2M) & 85 & 40$\rightarrow$\textbf{65} & 50$\rightarrow$\textbf{55} & 45$\rightarrow$40 & 75$\rightarrow$\textbf{90} & 80$\rightarrow$\textbf{85} & 80$\rightarrow$\textbf{100} & 85$\rightarrow$\textbf{90} & \textbf{$+$7.2}\,{\footnotesize $\pm$1.4} \\
 & BC-Transformer (18M) & 99 & 59$\rightarrow$\textbf{78} & 36$\rightarrow$\textbf{50} & 67$\rightarrow$61 & 93$\rightarrow$90 & 84$\rightarrow$\textbf{94} & 87$\rightarrow$\textbf{93} & 69$\rightarrow$\textbf{75} & \textbf{$+$6.8}\,{\footnotesize $\pm$0.2} \\
 & ACT~\cite{zhao2023act} ($\sim$52M) & 100 & 57$\rightarrow$\textbf{90} & 27$\rightarrow$\textbf{71} & 44$\rightarrow$\textbf{63} & 99$\rightarrow$\textbf{100} & 93$\rightarrow$\textbf{100} & 87$\rightarrow$\textbf{100} & 56$\rightarrow$\textbf{96} & \textbf{$+$22.2}\,{\footnotesize $\pm$1.4} \\
 & GR00T-N1.5~\cite{bjorck2025gr00t} (3B) & 100 & 40$\rightarrow$\textbf{78} & 13$\rightarrow$\textbf{54} & 17$\rightarrow$\textbf{81} & 93$\rightarrow$\textbf{99} & 61$\rightarrow$\textbf{100} & 49$\rightarrow$\textbf{97} & 16$\rightarrow$\textbf{92} & \textbf{$+$44.8}\,{\footnotesize $\pm$2.3} \\
 & OpenVLA~\cite{kim2025finetuning} (7B) & 100 & 50$\rightarrow$\textbf{65} & 10$\rightarrow$\textbf{55} & 40$\rightarrow$40 & 100$\rightarrow$100 & 96$\rightarrow$\textbf{98} & 100$\rightarrow$100 & 95$\rightarrow$\textbf{98} & \textbf{$+$9.3}\, {\footnotesize $\pm$2.5} \\

\midrule
\multirow{2}{*}{\rotatebox{90}{\textit{Can}}} & BC ($\sim$2M) & 39 & 9$\rightarrow$\textbf{11} & 2$\rightarrow$0 & 4$\rightarrow$0 & 13$\rightarrow$\textbf{35} & 15$\rightarrow$\textbf{24} & 16$\rightarrow$\textbf{43} & 3$\rightarrow$0 & \textbf{$+$7.3}\,{\footnotesize $\pm$2.9} \\
 & ACT ($\sim$52M) & 90 & 15$\rightarrow$\textbf{35} & 0$\rightarrow$0 & 10$\rightarrow$\textbf{20} & 85$\rightarrow$85 & 60$\rightarrow$\textbf{70} & 50$\rightarrow$50 & 0$\rightarrow$0 & \textbf{$+$6.2}\,{\footnotesize $\pm$1.6} \\
\midrule
\multirow{6}{*}{\rotatebox{90}{\textit{Square}}} & BC ($\sim$2M) & 34 & 1$\rightarrow$\textbf{7} & 0$\rightarrow$0 & 1$\rightarrow$0 & 24$\rightarrow$\textbf{33} & 12$\rightarrow$\textbf{30} & 9$\rightarrow$\textbf{29} & 2$\rightarrow$\textbf{11} & \textbf{$+$8.5}\,{\footnotesize $\pm$2.9} \\
 & BC-Transformer (18M) & 65 & 9$\rightarrow$\textbf{24} & 0$\rightarrow$\textbf{2} & 4$\rightarrow$\textbf{5} & 49$\rightarrow$\textbf{63} & 22$\rightarrow$\textbf{59} & 31$\rightarrow$\textbf{49} & 0$\rightarrow$\textbf{32} & \textbf{$+$17.1}\,{\footnotesize $\pm$0.3} \\
 & IRIS ($\sim$9M) & 74 & 3$\rightarrow$\textbf{17} & 1$\rightarrow$1 & 1$\rightarrow$\textbf{4} & 43$\rightarrow$\textbf{59} & 24$\rightarrow$\textbf{54} & 16$\rightarrow$\textbf{57} & 1$\rightarrow$\textbf{12} & \textbf{$+$16.6}\,{\footnotesize $\pm$0.6} \\
 & HBC ($\sim$12M) & 16 & 3$\rightarrow$\textbf{5} & 0$\rightarrow$0 & 0$\rightarrow$0 & 7$\rightarrow$\textbf{17} & 2$\rightarrow$\textbf{15} & 6$\rightarrow$\textbf{25} & 0$\rightarrow$\textbf{5} & \textbf{$+$7.0}\,{\footnotesize $\pm$2.6} \\
 & ACT ($\sim$52M) & 67 & 13$\rightarrow$\textbf{32} & 1$\rightarrow$\textbf{3} & 6$\rightarrow$5 & 69$\rightarrow$\textbf{71} & 51$\rightarrow$\textbf{65} & 41$\rightarrow$\textbf{64} & 0$\rightarrow$\textbf{27} & \textbf{$+$12.2}\,{\footnotesize $\pm$2.5} \\
\midrule
\multirow{3}{*}{\rotatebox{90}{\textit{Thread}}} & BC ($\sim$2M) & 73 & 8$\rightarrow$\textbf{29} & 1$\rightarrow$\textbf{4} & 1$\rightarrow$\textbf{3} & 41$\rightarrow$\textbf{54} & 33$\rightarrow$\textbf{57} & 34$\rightarrow$\textbf{49} & 0$\rightarrow$\textbf{53} & \textbf{$+$18.9}\,{\footnotesize $\pm$3.0} \\
 & BCQ ($\sim$6M) & 53 & 0$\rightarrow$\textbf{1} & 0$\rightarrow$\textbf{1} & 1$\rightarrow$0 & 15$\rightarrow$12 & 6$\rightarrow$\textbf{21} & 9$\rightarrow$\textbf{17} & 0$\rightarrow$\textbf{8} & \textbf{$+$4.3}\,{\footnotesize $\pm$1.2} \\
 & HBC ($\sim$12M) & 85 & 3$\rightarrow$\textbf{37} & 1$\rightarrow$\textbf{4} & 1$\rightarrow$\textbf{14} & 51$\rightarrow$\textbf{71} & 40$\rightarrow$\textbf{71} & 21$\rightarrow$\textbf{65} & 0$\rightarrow$\textbf{63} & \textbf{$+$29.7}\,{\footnotesize $\pm$4.6} \\
\midrule
\multirow{3}{*}{\rotatebox{90}{\textit{Stack}}} & BC ($\sim$2M) & 85 & 33$\rightarrow$\textbf{57} & 7$\rightarrow$\textbf{51} & 15$\rightarrow$\textbf{37} & 83$\rightarrow$83 & 67$\rightarrow$\textbf{83} & 58$\rightarrow$\textbf{79} & 4$\rightarrow$\textbf{75} & \textbf{$+$28.5}\,{\footnotesize $\pm$1.7} \\
 & BCQ ($\sim$6M) & 92 & 28$\rightarrow$\textbf{52} & 3$\rightarrow$1 & 11$\rightarrow$1 & 77$\rightarrow$\textbf{81} & 73$\rightarrow$\textbf{81} & 62$\rightarrow$\textbf{83} & 10$\rightarrow$\textbf{69} & \textbf{$+$14.8}\,{\footnotesize $\pm$2.9} \\
 & HBC ($\sim$12M) & 98 & 45$\rightarrow$\textbf{71} & 14$\rightarrow$8 & 20$\rightarrow$12 & 85$\rightarrow$\textbf{96} & 72$\rightarrow$\textbf{91} & 71$\rightarrow$\textbf{91} & 19$\rightarrow$\textbf{76} & \textbf{$+$16.8}\,{\footnotesize $\pm$2.8} \\
\bottomrule
\end{tabular}%
}
\end{table*}

\section{Experiments}
\label{sec:exp}

\begin{figure}[h]
    \centering
    \includegraphics[width=1\linewidth]{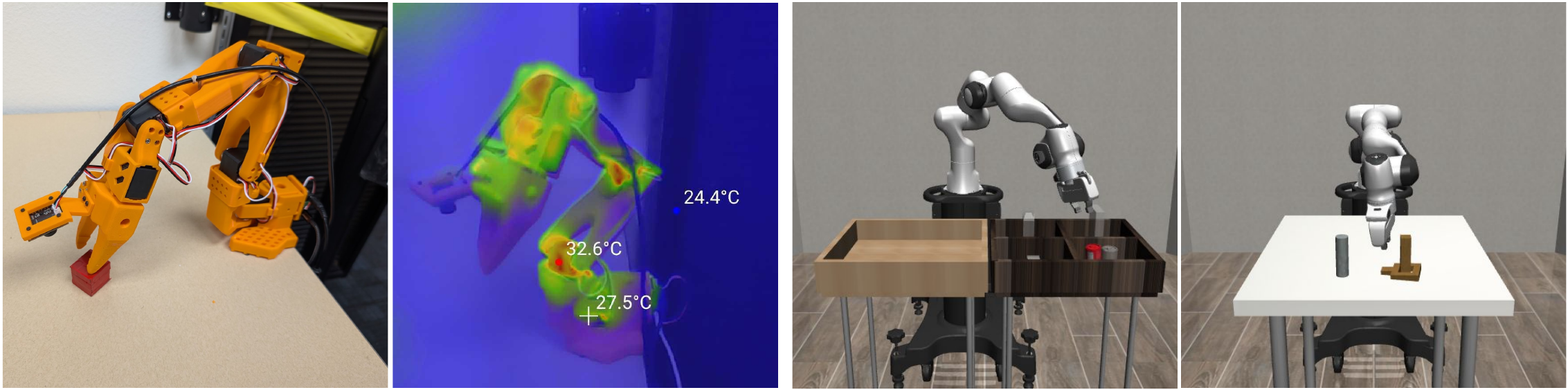}
    \caption{Environment setup for real (left-2) and sim (right-2).}
    \vspace{-0.5em}
    \label{fig:real_sim_examples}
\end{figure}

We organize the evaluation around five research questions. \textbf{RQ1. Effectiveness:} Does TeAR improve stressed-condition success across policy classes while preserving nominal behavior? \textbf{RQ2. Comparison against baselines:} How does it compare with robustness methods, alternative adapters, and training recipes? \textbf{RQ3. Architectural attribution:} Which components matter, and how does the trained adapter modify actions? \textbf{RQ4. Robustness to Degradation Mismatch:} Is TeAR effective when actuator behavior differs from the degradation model used for training? \textbf{RQ5. Sim-to-real transfer:} Does simulation-trained TeAR transfer to a physical arm?

\begin{figure}
    \centering
    \vspace{-1em}
    \includegraphics[width=1\linewidth]{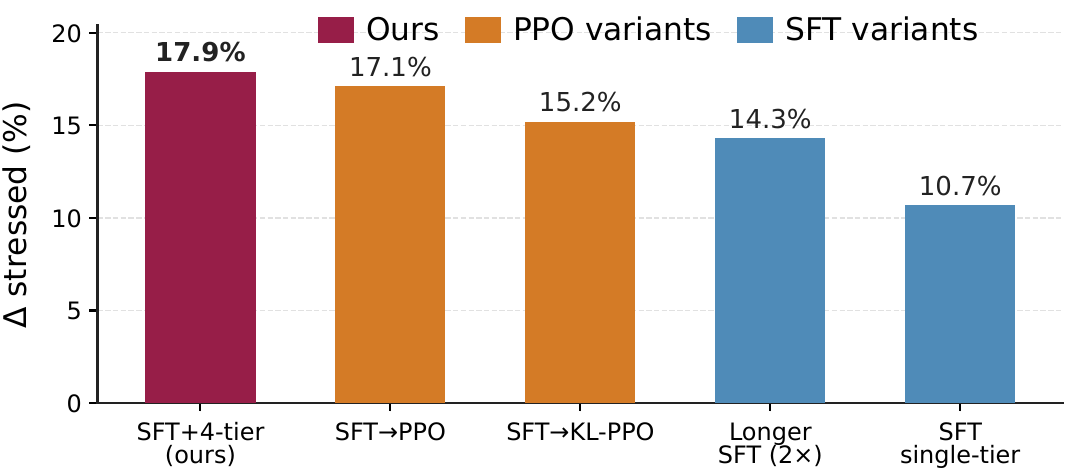}
    \caption{Stressed-condition $\Delta$ over the frozen base for five recipes with the TeAR architecture fixed.}
    \label{fig:rq2_recipes} 
    \vspace{-1em}
\end{figure}

\subsection{RQ1: Effectiveness}
\label{sec:rq1}

Table~\ref{tab:rq1_main} evaluates TeAR on 18 policy task pairs spanning eight policy families and five robosuite manipulation tasks~\cite{zhu2020robosuite,mandlekar2021robomimic}. The policies range from behavior cloning and hierarchical or offline-RL methods to ACT~\cite{zhao2023act} and VLA families~\cite{bjorck2025gr00t,kim2025finetuning}. TeAR achieves a positive mean gain across the seven stressed conditions for every evaluated pair, with the largest reported gain of $44.8\%$ on GR00T--Lift. These results show that telemetry-conditioned action correction is useful across different policy architectures, including large pre-trained models, without updating the base policy. The benefit nevertheless depends on the operating condition. For BC-Lift, TeAR increases success from 40\% to 65\% under heating, but decreases it from 45\% to 40\% under voltage stress. Thus, positive average gains do not imply uniform recovery across degradation channels. Under nominal telemetry, TeAR preserves the reported success rate for every evaluated pair, consistent with its structural pass-through gate. We separately verify exact action and trajectory identity in the paired evaluation (Section~\ref{sec:mismatch}).

\subsection{RQ2: Comparison against baselines}
\label{sec:rq2}

We systematically compare TeAR against three distinct families of alternatives below: robust-training baselines, alternative adapters, and alternative training recipes.

\textbf{Robust-training and adaptive-control baselines.} Table~\ref{tab:rq2_baselines} compares TeAR with robust-training and adaptive-control baselines. TeAR achieves the largest mean gain over the seven stressed conditions, $18.3\pm2.9$ percentage points, compared with $15.5\pm5.3$ for the MRAC/$L_1$ implementation. The advantage depends on the condition: MRAC/$L_1$ achieves higher success under hot and $\mathcal{TCV}_{\mathrm{mod}}$ stress, whereas TeAR performs better under $\mathcal{T}_{\mathrm{mod}}$ and $\mathcal{TC}_{\mathrm{mod}}$. The MRAC/$L_1$ implementation uses privileged simulator command-response feedback. Retraining-based alternatives yield smaller mean gains in this evaluation. Section~\ref{sec:mismatch} compares TeAR with direct compensation using an assumed degradation model.

\textbf{Alternative adapter architectures.} With the SFT recipe and gate fixed, Table~\ref{tab:rq2_adapter_archs} compares alternative adapter backbones. TeAR achieves a mean stressed gain of $18.3$ percentage points, compared with $6.0$ for cross-attention, the strongest alternative in this comparison. The separation is most apparent under combined temperature and current stress, where TeAR reaches $55.0\%$ and the alternative adapters reach at most $13.3\%$. These results support the chosen backbone among the configurations in this table; the separate MLP checkpoint in Table~\ref{tab:rq3_ablation} shows that the advantage does not hold across all evaluated variants.

\textbf{Alternative training recipes.}
Figure~\ref{fig:rq2_recipes} reports the original training-recipe evaluation, separately from the three-seed checkpoint comparisons. With the TeAR architecture fixed, four-tier SFT achieves the largest mean stressed-condition gain among the evaluated recipes: $17.9\%$, compared with $10.7\%$ for single-tier training. Adding PPO after SFT yields $17.1\%$, while the KL-constrained PPO variant reaches $15.2\%$. Doubling the SFT duration yields $14.3\%$. These results support four-tier supervised training for this configuration: neither the additional on-policy stage nor longer optimization improves the reported mean gain.

\begin{table}[t]
\centering
\vspace{0.5em}
\setlength{\tabcolsep}{2.4pt}
\caption{\textbf{Baselines.} Success (\%), mean $\pm$ sample SD across 3 evaluation seeds. $\Delta$ averages paired gains over the frozen BC-Transformer across seven stressed conditions.}
\label{tab:rq2_baselines}
\begin{tabular}{@{}lcccc r@{}}
\toprule
Method & hot & $\mathcal{T}_{\text{mod}}$ & $\mathcal{TC}_{\text{mod}}$ & $\mathcal{TCV}_{\text{mod}}$ & $\Delta(\uparrow)$ \\
\midrule
Frozen base & $6.7\pmsd{2.9}$ & $43.3\pmsd{12.6}$ & $26.7\pmsd{5.8}$ & $0.0\pmsd{0.0}$ & --- \\
\midrule
No-Tele. SFT & $3.3\pmsd{5.8}$ & $41.7\pmsd{10.4}$ & $0.0\pmsd{0.0}$ & $0.0\pmsd{0.0}$ & $-6.9\pmsd{3.4}$ \\
Domain R.~\cite{tobin2017domain} & $5.0\pmsd{0.0}$ & $46.7\pmsd{11.5}$ & $25.0\pmsd{13.2}$ & $0.0\pmsd{0.0}$ & $0.0\pmsd{3.1}$ \\
ACDR~\cite{okamoto2021reinforcement} & $8.3\pmsd{2.9}$ & $40.0\pmsd{10.0}$ & $31.7\pmsd{7.6}$ & $0.0\pmsd{0.0}$ & $+1.0\pmsd{3.2}$ \\
RMA-style~\cite{kumar2021rma} & $5.0\pmsd{5.0}$ & $35.0\pmsd{13.2}$ & $23.3\pmsd{5.8}$ & $0.0\pmsd{0.0}$ & $-4.3\pmsd{2.6}$ \\
BC fine-tune & $5.0\pmsd{0.0}$ & $46.7\pmsd{11.5}$ & $25.0\pmsd{13.2}$ & $0.0\pmsd{0.0}$ & $0.0\pmsd{3.1}$ \\
MRAC/$L_1$~\cite{hovakimyan2010l1} & $26.7\pmsd{10.4}$ & $50.0\pmsd{5.0}$ & $53.3\pmsd{18.9}$ & $30.0\pmsd{8.7}$ & $+15.5\pmsd{5.3}$ \\
\midrule
\textbf{TeAR} & $23.3\pmsd{2.9}$ & $63.3\pmsd{2.9}$ & $55.0\pmsd{5.0}$ & $21.7\pmsd{2.9}$ & $+18.3\pmsd{2.9}$ \\
\bottomrule
\end{tabular}
\end{table}

\begin{figure}[t]
\centering
\includegraphics[width=\linewidth]{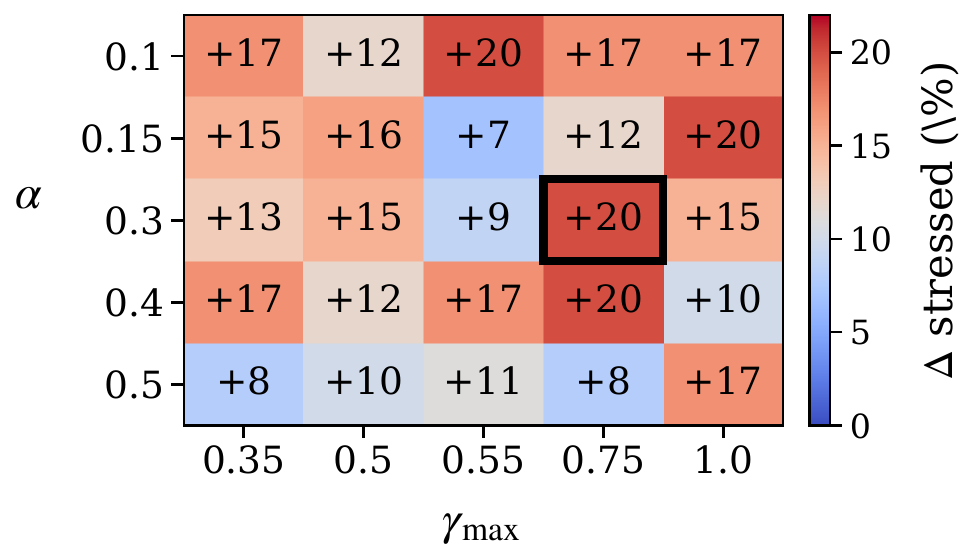}
\vspace{-1.8em}
\caption{Stressed success gain sweeping the additive-residual cap
$\alpha$ and the multiplicative gain $\gamma_{\max}$. The selected
$(\alpha,\gamma_{\max})=(0.30,0.75)$ is outlined.}
\label{fig:rq3_grid}
\vspace{-1em}
\end{figure}

\begin{table}[t]
\centering
\small
\vspace{0.5em}
\setlength{\tabcolsep}{1pt}
\renewcommand{\arraystretch}{1.05}
\caption{Success (\%) as mean $\pm$ sample SD across three evaluation seeds with fixed checkpoints. Stressed averages all seven stress conditions; $\Delta_{T,TV}$ is the paired gain over full TeAR averaged over $\mathcal{T}_{\text{mod}}$ and $\mathcal{TV}_{\text{mod}}$ conditions.}
\vspace{-0.4em}
\label{tab:rq3_ablation}
\begin{tabular}{@{}lcccc@{}}
\toprule
Configuration & $\mathcal{T}_{\text{mod}}(\uparrow)$ & $\mathcal{TV}_{\text{mod}}(\uparrow)$ & Stressed($\uparrow$) & $\Delta_{T,TV}$ \\
\midrule
\textbf{TeAR} (reference) & $63.3\pmsd{2.9}$ & $68.3\pmsd{12.6}$ & $34.8\pmsd{0.4}$ & $0.0\pmsd{0.0}$ \\
\midrule
Gate $\to$ smoothstep & $53.3\pmsd{10.4}$ & $48.3\pmsd{16.1}$ & $34.3\pmsd{3.1}$ & $-15.0\pmsd{13.2}$ \\
Output: $\delta$ only & $46.7\pmsd{5.8}$ & $58.3\pmsd{15.3}$ & $31.7\pmsd{4.2}$ & $-13.3\pmsd{5.2}$ \\
Longer SFT ($2\times$) & $61.7\pmsd{12.6}$ & $58.3\pmsd{2.9}$ & $33.8\pmsd{2.9}$ & $-5.8\pmsd{11.5}$ \\
Gate $\to$ linear & $55.0\pmsd{5.0}$ & $50.0\pmsd{10.0}$ & $33.1\pmsd{2.2}$ & $-13.3\pmsd{7.6}$ \\
Output: $\gamma$ only & $53.3\pmsd{11.5}$ & $51.7\pmsd{7.6}$ & $36.0\pmsd{3.9}$ & $-13.3\pmsd{12.6}$ \\
Scalar gate & $55.0\pmsd{5.0}$ & $53.3\pmsd{2.9}$ & $31.7\pmsd{2.7}$ & $-11.7\pmsd{8.0}$ \\
No state token & $36.7\pmsd{10.4}$ & $53.3\pmsd{14.4}$ & $27.1\pmsd{1.9}$ & $-20.8\pmsd{5.2}$ \\
MLP backbone & $61.7\pmsd{5.8}$ & $63.3\pmsd{5.8}$ & $41.0\pmsd{3.5}$ & $-3.3\pmsd{10.1}$ \\
\bottomrule
\end{tabular}
\end{table}

\subsection{RQ3: Architectural attribution and correction behavior}
\label{sec:rq3}

Table~\ref{tab:rq3_ablation} examines variants of the correction heads, state input, gate, and backbone. Removing the state token reduces mean stressed success from $34.8\%$ to $27.1\%$. The residual-only and gain-only variants achieve $31.7\%$ and $36.0\%$, respectively, indicating that both heads are not essential for improving performance in this setting. The MLP checkpoint achieves the highest overall stressed mean at $41.0\%$, while full TeAR achieves the highest means on $\mathcal{T}_{\mathrm{mod}}$ and $\mathcal{TV}_{\mathrm{mod}}$. The relative performance therefore depends on the stress condition. The $\Delta_{T,TV}$ column summarizes these two displayed conditions; Stressed averages all seven.

The gate variants also change gain parameterization, action-magnitude input, and gripper masking, so they do not isolate gate shape. The smoothstep variant retains the reference gate shape. The sensitivity sweep in Fig.~\ref{fig:rq3_grid} motivates the deployment bounds $(\alpha,\gamma_{\max})=(0.30,0.75)$, which we use throughout evaluation.

\begin{figure*}[t]
\centering
\vspace{0.5em}
\includegraphics[width=\textwidth]{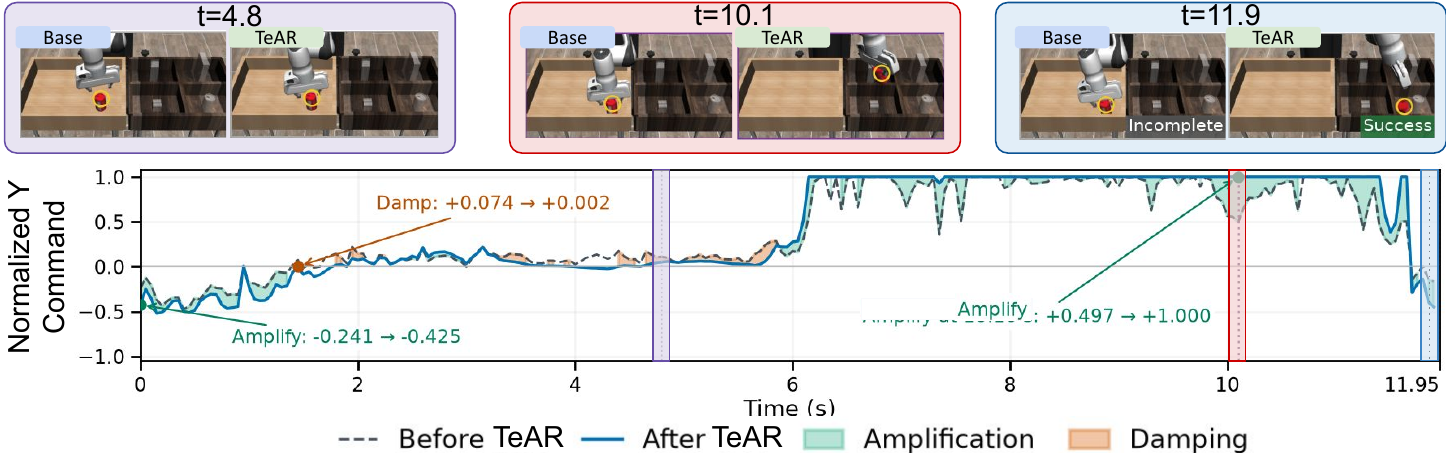}
\caption{Action correction under combined temperature and current stress. Top: paired base-policy and TeAR snapshots from the same initial state; yellow circles mark the target can. TeAR completes the task at 11.95\,s, while the base fails within 20\,s. Bottom: proposed and corrected $y$-translation commands along the TeAR trajectory. Shading highlights amplification and damping, with numeric examples of each. This selected rollout illustrates correction behavior.}
\vspace{-1em}
\label{fig:rq3_corrections}
\end{figure*}

\textbf{Correction behavior.} Figure~\ref{fig:rq3_corrections} shows action correction during a Can rollout under combined temperature and current stress. From the same initial scene, TeAR completes the placement while the base policy fails within the evaluation horizon. Along the TeAR trajectory, the adapter both damps commands from $0.074$ to $0.002$ and amplifies them from $0.497$ to $1.000$. This selected rollout illustrates correction behavior; aggregate evaluations measure task success.

\begin{table}[t]
\vspace{-1em}
\centering
\setlength{\tabcolsep}{2.8pt}
\caption{\textbf{Adapter architectures.} Success (\%), mean $\pm$ sample SD across three evaluation seeds with fixed checkpoints, using the same SFT recipe and gate. $\Delta$ averages gains over the frozen base across seven stressed conditions.}
\label{tab:rq2_adapter_archs}
\begin{tabular}{@{}lcccc r@{}}
\toprule
Adapter & hot & $\mathcal{T}_{\text{mod}}$ & $\mathcal{TC}_{\text{mod}}$ & $\mathcal{TCV}_{\text{mod}}$ & $\Delta (\uparrow)$ \\
\midrule
Frozen BC-T & $6.7\pmsd{2.9}$ & $43.3\pmsd{12.6}$ & $26.7\pmsd{5.8}$ & $0.0\pmsd{0.0}$ & --- \\
Bottleneck res & $10.0\pmsd{0.0}$ & $51.7\pmsd{11.5}$ & $1.7\pmsd{2.9}$ & $10.0\pmsd{5.0}$ & $+1.9\pmsd{3.5}$ \\
Cross-attn & $15.0\pmsd{10.0}$ & $55.0\pmsd{5.0}$ & $10.0\pmsd{5.0}$ & $13.3\pmsd{10.4}$ & $+6.0\pmsd{6.8}$ \\
FiLM & $28.3\pmsd{2.9}$ & $20.0\pmsd{5.0}$ & $13.3\pmsd{5.8}$ & $8.3\pmsd{5.8}$ & $+1.0\pmsd{2.3}$ \\
Per-joint MLP & $15.0\pmsd{5.0}$ & $45.0\pmsd{13.2}$ & $3.3\pmsd{2.9}$ & $6.7\pmsd{5.8}$ & $0.0\pmsd{5.0}$ \\
Global MLP & $23.3\pmsd{2.9}$ & $36.7\pmsd{16.1}$ & $8.3\pmsd{7.6}$ & $10.0\pmsd{0.0}$ & $+1.2\pmsd{5.8}$ \\
Hybrid & $20.0\pmsd{8.7}$ & $63.3\pmsd{2.9}$ & $6.7\pmsd{7.6}$ & $8.3\pmsd{5.8}$ & $+4.5\pmsd{5.4}$ \\
MoE & $18.3\pmsd{2.9}$ & $61.7\pmsd{12.6}$ & $3.3\pmsd{5.8}$ & $10.0\pmsd{10.0}$ & $+4.5\pmsd{3.2}$ \\
\midrule
\textbf{TeAR} & $23.3\pmsd{2.9}$ & $63.3\pmsd{2.9}$ & $55.0\pmsd{5.0}$ & $21.7\pmsd{2.9}$ & $+18.3\pmsd{2.9}$ \\
\bottomrule
\end{tabular}
\vspace{-1.5em}
\end{table}

\begin{figure}[t]
\centering
\includegraphics[width=\linewidth]{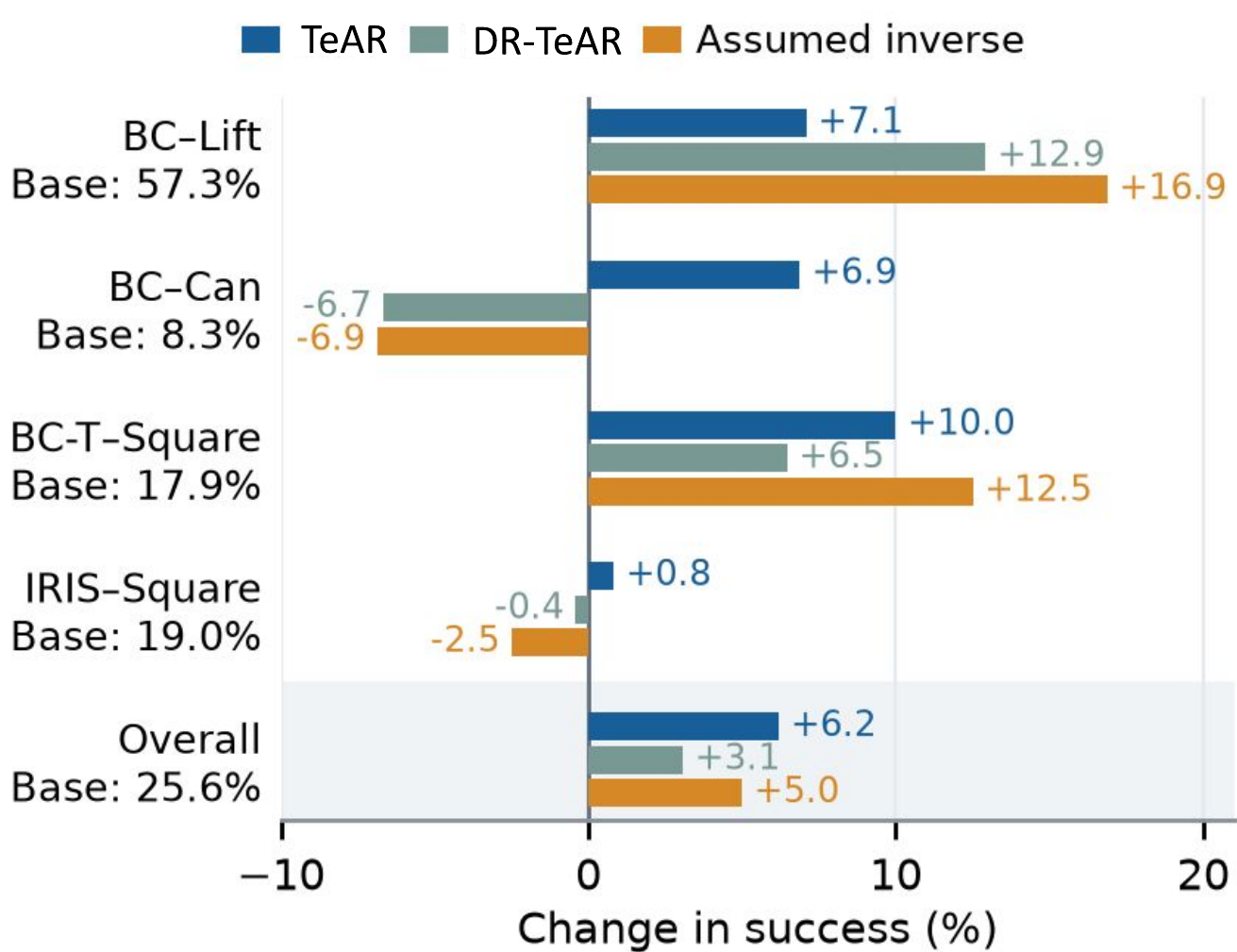}
\caption{Compensation under model mismatch. Bars show corrected minus base success rates; positive values indicate improvement. Each task uses 480 episodes per method; Overall averages the four tasks.}
\label{fig:rq4_success}
\vspace{-1em}
\end{figure}

\subsection{RQ4: Robustness to degradation-model mismatch}
\label{sec:mismatch}

Actuator response at deployment may differ from the model used to train TeAR. We test whether its correction remains effective under this mismatch and whether randomized-curve training improves transfer. We evaluate eight temperature--current--voltage curve triples drawn from linear, exponential, sigmoid, and polynomial families; each includes a nonlinear channel absent from TeAR's nominal linear training model. The evaluation covers BC (Lift, Can), BC-Transformer (Square), and IRIS (Square) under four stress profiles (hot, $\mathcal T_{\mathrm{mod}}$, $\mathcal{TC}_{\mathrm{mod}}$, and $\mathcal{TCV}_{\mathrm{mod}}$). Five episodes per profile and three seeds give 480 stressed episodes per pair and method, with initial states, telemetry, and random streams paired across methods and all checkpoints and correction bounds fixed.

We compare TeAR with an assumed-model inverse that compensates commands using the nominal training curves, $a'_d=\mathrm{clip}(a_{b,d}/q_d,-1,1),$ where $q_d=q_p$ for translation and $q_d=q_r$ for rotation (Eq.~\ref{eq:grouped}); the gripper is unchanged. Like TeAR, this inverse has no access to the actual test capacities. We also evaluate DR-TeAR, trained with 64 randomized curve triples using the same supervised procedure. Its test curves use fresh parameters and combinations from families encountered during training. This is a comparison of existing trained variants: the Can checkpoints differ in action-input noise, and the Lift checkpoints differ in training base-policy metadata, so their differences cannot be attributed to curve randomization alone.

TeAR raises mean success under model mismatch from $25.6\%$ to $31.8\%$, with positive mean gains on all four policy--task pairs (Fig.~\ref{fig:rq4_success}). The assumed-model inverse reaches $30.6\%$; its gains on Lift and BC-Transformer--Square are accompanied by regressions on Can and IRIS--Square. Randomized-curve training also improves on the base, with DR-TeAR reaching $28.7\%$, although it does not consistently improve on TeAR. Across response models, TeAR improves success on six of eight curve draws. The paired gains over the base and assumed inverse are $6.2$ and $1.2$ percentage points, with 95\% curve-bootstrap intervals of $[-0.1,11.2]$ and $[-3.3,6.1]$, respectively. These intervals resample whole curve draws jointly across tasks and include zero: the observed gains suggest transfer beyond the training model, but do not establish a conclusive aggregate advantage across response models. TeAR preserves exact nominal actions and trajectories in all 60 paired episodes.

\begin{table}
\centering
\small
\setlength{\tabcolsep}{9pt}
\caption{\textbf{SO-101 sim-to-real.} Lift success rates (\%) over 20 trials per method and condition; $\Delta$ is the \% gain.}
\label{tab:rq5_hardware}
\begin{tabular}{lccc}
\toprule
Condition & Base & TeAR & $\Delta (\uparrow)$ \\
\midrule
Cool ($<45^\circ$C)              & 100 & 100 & $\pm 0$ \\
Hot (shoulder $>50^\circ$C) & 75 & 85 & $+10$ \\
Hot (shoulder \& wrist $>50^\circ$C)        & 70 & 85 & $+15$ \\
\bottomrule
\vspace{-2em}
\end{tabular}
\end{table}

\subsection{RQ5: Sim-to-real transfer}
\label{sec:rq5}

To test transfer beyond simulation, we deploy TeAR on a low-cost SO-ARM101~\cite{knight2024soarm} (SO-101) five-DoF arm performing the Lift task (Fig.~\ref{fig:real_sim_examples}). The base policy is behavior-cloned in simulation, and the adapter is trained using the simulation-derived compensation targets in Eq.~\ref{eq:sftarget}. Both are deployed without on-robot fine-tuning. We evaluate three operating conditions: cool operation below $45^\circ$C, shoulder heating above $50^\circ$C, and heating of both the shoulder and wrist above $50^\circ$C. The heated conditions are induced by operating the arm under load. Table~\ref{tab:rq5_hardware} shows that TeAR maintains 85\% success in both heated conditions, compared with 75\% and 70\% for the frozen base. These gains of 10 and 15 percentage points correspond to two and three additional successful trials. As heating extends from the shoulder to the wrist, base-policy success decreases while TeAR's success remains unchanged. Under cool conditions, both methods achieve 100\% success. The results therefore provide task-level evidence that a simulation-trained correction can improve execution on physical hardware under heating without adapting the base policy. This evaluation quantifies thermal transfer on one task; it does not separately measure the effects of physical current or voltage variation. Equal cool-condition success indicates preserved task performance, while exact action identity is established by the gate and the separate nominal checks in simulation.

\section{Conclusion}
\label{sec:conclusion}

We introduced Telemetry-Aware Action Rectification (TeAR), which uses onboard temperature, current, and voltage measurements to correct a frozen manipulation policy's commands before execution. Trained in simulation with physics-derived supervision, the adapter learns multiplicative and additive corrections while preserving the base command whenever telemetry lies within the nominal bands. Evaluation across 18 policy--task pairs, eight policy families, and five manipulation tasks demonstrates the usefulness of this interface across diverse policies. The additional paired evaluation shows positive mean gains under changes to the degradation curves, although recovery varies across tasks and response models. Randomized-curve training also improves on the frozen base, but does not consistently outperform the original TeAR. On a physical SO-101 arm, the simulation-trained adapter raises success from 75\% and 70\% to 85\% under two thermal-stress conditions without on-robot fine-tuning. Together, these results support onboard telemetry as a useful source of deployment-time context for improving manipulation under changing actuator conditions. Future work will investigate learning telemetry response relationships from operational data and evaluating transfer across robots under combined thermal, current, and voltage stress.

\bibliographystyle{unsrt}   
\bibliography{example}         

@article{black2026real,
  title={Real-time execution of action chunking flow policies},
  author={Black, Kevin and Galliker, Manuel and Levine, Sergey},
  journal={Advances in Neural Information Processing Systems},
  volume={38},
  pages={33383--33407},
  year={2026}
}

@article{rugh2000research,
  title={Research on gain scheduling},
  author={Rugh, Wilson J and Shamma, Jeff S},
  journal={Automatica},
  volume={36},
  number={10},
  pages={1401--1425},
  year={2000},
  publisher={Elsevier}
}

@article{youn2024thermal,
  title={Thermal characteristic modeling and compensation for the improvement of actuator homeostasis},
  author={Youn, Jimin and Kim, Hyeongjun and Shi, Kyeongsu and Kong, Kyoungchul},
  journal={IEEE/ASME Transactions on Mechatronics},
  volume={29},
  number={4},
  pages={3019--3027},
  year={2024},
  publisher={IEEE}
}

@article{silver2018residual,
  title={Residual policy learning},
  author={Silver, Tom and Allen, Kelsey and Tenenbaum, Josh and Kaelbling, Leslie},
  journal={arXiv preprint arXiv:1812.06298},
  year={2018}
}

@inproceedings{johannink2019residual,
  title={Residual reinforcement learning for robot control},
  author={Johannink, Tobias and Bahl, Shikhar and Nair, Ashvin and Luo, Jianlan and Kumar, Avinash and Loskyll, Matthias and Ojea, Juan Aparicio and Solowjow, Eugen and Levine, Sergey},
  booktitle={2019 international conference on robotics and automation (ICRA)},
  pages={6023--6029},
  year={2019},
  organization={IEEE}
}

@inproceedings{ankile2024imitation,
  title={From imitation to refinement--residual rl for precise visual assembly},
  author={Ankile, Lars Lien and Simeonov, Anthony and Shenfeld, Idan and Villasevil, Marcel Torne and Agrawal, Pulkit},
  booktitle={CoRL 2024 Workshop on Mastering Robot Manipulation in a World of Abundant Data},
  year={2024}
}

@inproceedings{yuan2024policy,
  title={Policy decorator: Model-agnostic online refinement for large policy model},
  author={Yuan, Xiu and Mu, Tongzhou and Tao, Stone and Fang, Yunhao and Zhang, Zhang and Su, Hao},
  booktitle={International Conference on Learning Representations},
  volume={2025},
  pages={28129--28164},
  year={2025}
}

@article{wagenmaker2025steering,
  title={Steering your diffusion policy with latent space reinforcement learning},
  author={Wagenmaker, Andrew and Nakamoto, Mitsuhiko and Zhang, Yunchu and Park, Seohong and Yagoub, Waleed and Nagabandi, Anusha and Gupta, Abhishek and Levine, Sergey},
  journal={arXiv preprint arXiv:2506.15799},
  year={2025}
}

@article{kim2024openvla,
  title={Openvla: An open-source vision-language-action model},
  author={Kim, Moo Jin and Pertsch, Karl and Karamcheti, Siddharth and Xiao, Ted and Balakrishna, Ashwin and Nair, Suraj and Rafailov, Rafael and Foster, Ethan and Lam, Grace and Sanketi, Pannag and others},
  journal={arXiv preprint arXiv:2406.09246},
  year={2024}
}

@article{kim2025finetuning,
  title={Fine-tuning vision-language-action models: Optimizing speed and success},
  author={Kim, Moo Jin and Finn, Chelsea and Liang, Percy},
  journal={arXiv preprint arXiv:2502.19645},
  year={2025}
}

@article{black2024pi0,
  title={{$\pi_0$}: A Vision-Language-Action Flow Model for General Robot Control},
  author={Black, Kevin and Brown, Noah and Driess, Danny and Esmail, Adnan and Equi, Michael and Finn, Chelsea and Fusai, Niccolo and Groom, Lachy and Hausman, Karol and Ichter, Brian and others},
  journal={arXiv preprint arXiv:2410.24164},
  year={2024}
}

@article{octo2024,
  title={Octo: An open-source generalist robot policy},
  author={Team, Octo Model and Ghosh, Dibya and Walke, Homer and Pertsch, Karl and Black, Kevin and Mees, Oier and Dasari, Sudeep and Hejna, Joey and Kreiman, Tobias and Xu, Charles and others},
  journal={arXiv preprint arXiv:2405.12213},
  year={2024}
}

@inproceedings{jiang2024transic,
  title     = {TRANSIC: Sim-to-Real Policy Transfer by Learning from Online Correction},
  author    = {Yunfan Jiang and Chen Wang and Ruohan Zhang and Jiajun Wu and Li Fei-Fei},
  booktitle = {Conference on Robot Learning},
  year      = {2024}
}

@inproceedings{kumar2021rma,
title={Rma: Rapid motor adaptation for legged robots},
author={Kumar, Ashish and Fu, Zipeng and Pathak, Deepak and Malik, Jitendra},
booktitle={Robotics: Science and Systems},
year={2021}
}

@article{yu2017upops,
  title={Preparing for the unknown: Learning a universal policy with online system identification},
  author={Yu, Wenhao and Tan, Jie and Liu, C Karen and Turk, Greg},
  journal={arXiv preprint arXiv:1702.02453},
  year={2017}
}

@article{nagabandi2019meta,
  title={Learning to adapt in dynamic, real-world environments through meta-reinforcement learning},
  author={Nagabandi, Anusha and Clavera, Ignasi and Liu, Simin and Fearing, Ronald S and Abbeel, Pieter and Levine, Sergey and Finn, Chelsea},
  journal={arXiv preprint arXiv:1803.11347},
  year={2018}
}

@inproceedings{tobin2017domain,
  title={Domain randomization for transferring deep neural networks from simulation to the real world},
  author={Tobin, Josh and Fong, Rachel and Ray, Alex and Schneider, Jonas and Zaremba, Wojciech and Abbeel, Pieter},
  booktitle={2017 IEEE/RSJ international conference on intelligent robots and systems (IROS)},
  pages={23--30},
  year={2017},
  organization={IEEE}
}

@inproceedings{peng2018simtoreal,
  title={Sim-to-real transfer of robotic control with dynamics randomization},
  author={Peng, Xue Bin and Andrychowicz, Marcin and Zaremba, Wojciech and Abbeel, Pieter},
  booktitle={2018 IEEE international conference on robotics and automation (ICRA)},
  pages={3803--3810},
  year={2018},
  organization={IEEE}
}

@article{akkaya2019solving,
  title={Solving rubik's cube with a robot hand},
  author={Akkaya, Ilge and Andrychowicz, Marcin and Chociej, Maciek and Litwin, Mateusz and McGrew, Bob and Petron, Arthur and Paino, Alex and Plappert, Matthias and Powell, Glenn and Ribas, Raphael and others},
  journal={arXiv preprint arXiv:1910.07113},
  year={2019}
}

@inproceedings{pinto2017robust,
  title={Robust adversarial reinforcement learning},
  author={Pinto, Lerrel and Davidson, James and Sukthankar, Rahul and Gupta, Abhinav},
  booktitle={International conference on machine learning},
  pages={2817--2826},
  year={2017},
  organization={PMLR}
}

@article{okamoto2021reinforcement,
  title={Reinforcement learning with adaptive curriculum dynamics randomization for fault-tolerant robot control},
  author={Okamoto, Wataru and Kera, Hiroshi and Kawamoto, Kazuhiko},
  journal={arXiv preprint arXiv:2111.10005},
  year={2021}
}

@article{khatib1987osc,
  title={A unified approach for motion and force control of robot manipulators: The operational space formulation},
  author={Khatib, Oussama},
  journal={IEEE Journal on Robotics and Automation},
  volume={3},
  number={1},
  pages={43--53},
  year={1987},
  publisher={IEEE}
}

@article{zhao2023act,
  title={Learning fine-grained bimanual manipulation with low-cost hardware},
  author={Zhao, Tony Z and Kumar, Vikash and Levine, Sergey and Finn, Chelsea},
  journal={arXiv preprint arXiv:2304.13705},
  year={2023}
}

@article{bjorck2025gr00t,
  title={Gr00t n1: An open foundation model for generalist humanoid robots},
  author={Bjorck, Johan and Casta{\~n}eda, Fernando and Cherniadev, Nikita and Da, Xingye and Ding, Runyu and Fan, Linxi and Fang, Yu and Fox, Dieter and Hu, Fengyuan and Huang, Spencer and others},
  journal={arXiv preprint arXiv:2503.14734},
  year={2025}
}

@article{zhu2020robosuite,
  title={robosuite: A modular simulation framework and benchmark for robot learning},
  author={Zhu, Yuke and Wong, Josiah and Mandlekar, Ajay and Mart{\'\i}n-Mart{\'\i}n, Roberto and Joshi, Abhishek and Lin, Kevin and Maddukuri, Abhiram and Nasiriany, Soroush and Zhu, Yifeng},
  journal={arXiv preprint arXiv:2009.12293},
  year={2020}
}

@article{mandlekar2021robomimic,
  title={What matters in learning from offline human demonstrations for robot manipulation},
  author={Mandlekar, Ajay and Xu, Danfei and Wong, Josiah and Nasiriany, Soroush and Wang, Chen and Kulkarni, Rohun and Fei-Fei, Li and Savarese, Silvio and Zhu, Yuke and Mart{\'\i}n-Mart{\'\i}n, Roberto},
  journal={arXiv preprint arXiv:2108.03298},
  year={2021}
}

@book{hovakimyan2010l1,
  title={L1 adaptive control theory: Guaranteed robustness with fast adaptation},
  author={Hovakimyan, Naira and Cao, Chengyu},
  year={2010},
  publisher={SIAM}
}

@misc{knight2024soarm,
  author       = {Rob Knight and Pepijn Kooijmans and Remi Cadene and
                  Simon Alibert and Michel Aractingi and Dana Aubakirova and Adil Zouitine and Russi Martino and Steven Palma and
                  Caroline Pascal and Thomas Wolf},
  title        = {Standard Open {SO-100} \& {SO-101} Arms},
  year         = {2024},
  howpublished = {https://github.com/TheRobotStudio/SO-ARM100},
}
\clearpage
\appendix

\subsection{Degradation model: per-channel forms}
\label{app:degradation}

This appendix describes the degradation model used to generate training targets and simulate actuator stress in Section~\ref{sec:degradation}. Temperature is measured in degrees Celsius; current and voltage are normalized ratios. Capacity decreases linearly with stress, while a separate smoothstep gate controls when TeAR applies a correction.

\textbf{Capacity factors.} Define the clipped severity coordinates
\begin{align}
u_T(T)&=\mathrm{clip}\!\left(\frac{T-43}{32},0,1\right),\\
u_C(C)&=\mathrm{clip}\!\left(\frac{C-0.60}{0.40},0,1\right),\\
u_V(V)&=\mathrm{clip}\!\left(\frac{0.90-V}{0.40},0,1\right).
\end{align}
The corresponding capacity factors are
\begin{equation}
\rho_c(x)=1-(1-\rho_{c,\min})u_c(x),
\qquad c\in\{T,C,V\},
\label{eq:app_capacity}
\end{equation}
with $(\rho_{T,\min},\rho_{C,\min},\rho_{V,\min}) =(0.05,0.10,0.10)$. Thus temperature capacity decreases between $43$ and $75^\circ$C, current capacity decreases between $0.60$ and $1.00$, and voltage capacity decreases as voltage falls from $0.90$ to $0.50$. These curves provide repeatable simulation conditions and are not fitted to a specific physical actuator.

\textbf{Applying degradation to motion commands.} We apply degradation to operational-space controller (OSC) commands using translation and rotation groups $G_p=\{1,2,3,4\}$ and $G_r=\{4,5,6,7\}$, with joint indices starting at one. For each channel,
\begin{equation}
\bar\rho_{c,k}=\frac{1}{|G_k|}\sum_{j\in G_k}\rho_c(x_{c,j}),
\qquad q_k=\prod_{c\in\{T,C,V\}}\bar\rho_{c,k}.
\end{equation}
The three translation coordinates share $q_p$ and the three rotation coordinates share $q_r$. The degradation wrapper leaves the gripper command unchanged. The model acts on Cartesian motion commands supplied to the controller, rather than directly on joint torques.

\textbf{Noise, ripple, and channel composition.} In addition to capacity scaling, temperature and voltage introduce noise, while current introduces ripple. As in Section~\ref{sec:degradation}, the wrapper applies temperature, current, and voltage effects sequentially:
\begin{equation}
a_{\mathrm{exec}}=\mathcal V\!\left(\mathcal C\!\left(\mathcal T(a_f)\right)\right).
\end{equation}
Commands are clipped to $[-1,1]$ after each channel, and the perturbations vanish at unit capacity. The supervised targets below compensate for deterministic capacity loss and omit these perturbations.

\textbf{Supervised compensation targets.} The SFT target uses index-wise telemetry products for the six arm coordinates,
\begin{equation}
\widetilde\rho_j=
\begin{cases}
\rho_T(T_j)\rho_C(C_j)\rho_V(V_j),&1\leq j\leq6,\\
1,&j=7.
\end{cases}
\end{equation}
For demonstration action $a_d$, the target is
\begin{equation}
a^\star_j=\mathrm{clip}\!\left(
\frac{a_{d,j}}{\max(\widetilde\rho_j,0.05)},-1,1\right).
\label{eq:app_target}
\end{equation}
This gives a training target directly from each demonstration, without running the environment. The denominator is bounded below by $0.05$, the action stays within $[-1,1]$, and the gripper command is preserved. Each arm target uses its corresponding telemetry entry, providing a deterministic approximation to the grouped simulation model.

When actuator response has been characterized, the measured capacity curves can be used directly to construct the supervised targets. When the response is uncertain, training can instead sample a range of plausible capacity curves, as in DR-TeAR, to cover variation in how telemetry relates to action execution. In both cases, deployment uses the measured telemetry without requiring access to the true capacity curves.

\begin{figure}[t]
\centering
\includegraphics[width=\linewidth]{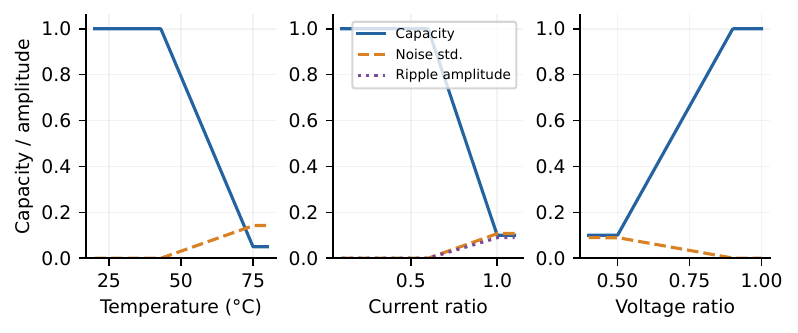}
\caption{\textbf{Per-channel capacity and perturbation magnitudes.} The implemented capacities are linear ramps. Dashed lines show Gaussian standard deviations; the dotted current trace shows ripple amplitude. All noise and ripple terms vanish at full capacity.}
\label{fig:degdetail}
\end{figure}

\subsection{Hyperparameters and training details}
\label{app:hp}

\textbf{Architecture.} The reference adapter uses three Transformer encoder layers, model width $128$, four attention heads, feed-forward width $256$, GELU activation, pre-layer normalization, and zero dropout. Seven telemetry--action tokens and one projected robot-state token form the input sequence. Learned positional embeddings distinguish token indices. A final layer normalization precedes the gain and residual heads, which are initialized to zero. Self-attention lets each action token use telemetry from all joints.

\textbf{Gate parameters.} With $h(z)=z^2(3-2z)$, the per-index gate is
\begin{align}
z_{T,j}&=\mathrm{clip}((T_j-42)/13,0,1),\nonumber\\
z_{C,j}&=\mathrm{clip}((C_j-0.60)/0.40,0,1),\nonumber\\
z_{V,j}&=\mathrm{clip}((0.90-V_j)/0.40,0,1),\\
g_j&=\min\{1,h(z_{T,j})+h(z_{C,j})+h(z_{V,j})\}.
\end{align}
Stress in any channel can activate TeAR. The gate sets when correction begins, while the capacity curves set the simulated loss of actuation. The temperature gate transitions from $42$ to $55^\circ$C, whereas the nominal temperature capacity ramp extends from $43$ to $75^\circ$C.

\begin{figure}[t]
\centering
\includegraphics[width=\linewidth]{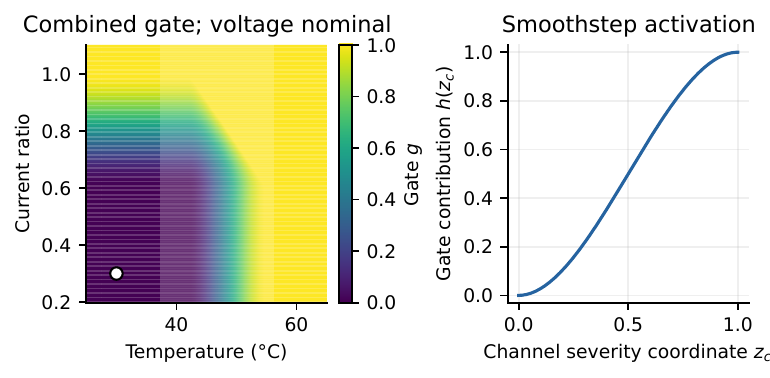}
\caption{\textbf{Telemetry gate.} Left: the combined temperature--current gate with nominal voltage; the marker lies in the exactly closed nominal region. Right: each channel uses the same smoothstep in its own normalized severity coordinate. Temperature gate corners are $42$--$55^\circ$C; current and voltage corners are $0.60$--$1.00$ and $0.90$--$0.50$. Gate activation is distinct from the capacity ramps in Fig.~\ref{fig:degdetail}.}
\label{fig:gate}
\end{figure}

\textbf{Exact pass-through under nominal telemetry.} For finite network outputs and normalized base commands, $T_j\leq42$, $C_j\leq0.60$, and $V_j\geq0.90$ imply $g_j=0$. Consequently $\gamma_j=1$, $\delta_j=0$, and $a_{f,j}=a_{b,j}$. The gate therefore preserves the base command exactly under nominal telemetry, independently of the learned weights.

\textbf{Bounded correction.} At deployment we use $\alpha=0.30$ and $\gamma_{\max}=0.75$. The output parameterization gives
\begin{equation}
|a_{f,j}-a_{b,j}|
\leq g_j\bigl(\gamma_{\max}|a_{b,j}|+\alpha\bigr).
\label{eq:app_correction_bound}
\end{equation}
This follows because $|\tanh(\cdot)|\leq1$ and clipping cannot increase the distance from a base command in $[-1,1]$. The gate therefore controls both when TeAR acts and how much it can change each command. This is a bound on the command correction, not a stability guarantee.

\textbf{Adapter size.} Table~\ref{tab:rq3_tam_family} compares adapter sizes on three policy--task combinations: BC-Transformer--Can, BC--Threading, and BC--Stack, with seed $42$ and $20$ episodes per condition. The mean covers the seven stress conditions in Appendix~\ref{app:severity}. The compact reference adapter matches the larger model in average success, supporting its use on these tasks.
\begin{table}[t]
\centering\footnotesize
\caption{\textbf{Adapter size comparison.} Success (\%) under the same seven-condition aggregation.}
\label{tab:rq3_tam_family}
\begin{tabular}{lrrrr}\toprule
Trunk & Can & Threading & Stack & Mean\\\midrule
Reference & $12.9$ & $35.0$ & $62.9$ & $36.9$\\
XL & $13.6$ & $30.7$ & $66.4$ & $36.9$\\
\bottomrule\end{tabular}
\end{table}

\subsection{Supervised training and telemetry curriculum}
\label{app:training}

\textbf{Training samples.} Each demonstration state--action pair is augmented with one telemetry profile from each of the four tiers in Table~\ref{tab:app_tiers}. Channel values are sampled uniformly within each tier's ranges for each telemetry index. This produces four supervised examples per demonstration time step, without collecting adapter rollouts.

\begin{table}[t]
\centering\footnotesize
\setlength{\tabcolsep}{4pt}
\caption{\textbf{Telemetry ranges for SFT augmentation.} Current and voltage are normalized ratios. Each tier contributes equally many examples.}
\label{tab:app_tiers}
\begin{tabular}{lccc}
\toprule
Tier & Temperature ($^\circ$C) & Current & Voltage\\
\midrule
Clean & $[20,42]$ & $[0.10,0.50]$ & $[0.92,1.00]$\\
Mild & $[43,55]$ & $[0.50,0.75]$ & $[0.80,0.92]$\\
Moderate & $[50,65]$ & $[0.65,0.90]$ & $[0.65,0.85]$\\
Severe & $[60,75]$ & $[0.80,1.05]$ & $[0.45,0.70]$\\
\bottomrule
\end{tabular}
\end{table}

The SFT implementation supplies the demonstration action as the adapter's action input and Eq.~\ref{eq:app_target} as the target. At deployment, this input is replaced by the frozen policy's proposed action. Where enabled, action-input augmentation adds Gaussian noise and clips the input to $[-1,1]$; the target remains derived from the original demonstration action.

\textbf{Objective and optimization.} Training minimizes the mean absolute target error plus three times the mean absolute identity error on clean samples. We use AdamW with initial learning rate $5\times10^{-4}$, weight decay $10^{-4}$, default Adam betas $(0.9,0.999)$, and $\epsilon=10^{-8}$. The learning rate follows cosine annealing to zero over the configured SFT duration, without a warm-up stage. Minibatches contain $256$ augmented state--action samples, and gradient norms are clipped to $1.0$.

The four reference TeAR checkpoints used in the paired mismatch study were trained for $30{,}000$ SFT steps with training seed $42$. They used $\alpha=0.30$ and gain-deviation bound $0.50$ during training, with the fixed deployment bound $0.75$ for evaluation. The BC--Can checkpoint used action-input noise with standard deviation $0.15$; the other three reference checkpoints used unperturbed demonstration-action inputs. These settings apply to the saved checkpoints used in the paired mismatch study; training durations vary across experiments.

\subsection{SO-101 hardware setup}
\label{app:so101}
\begin{figure}[t]
\centering
\includegraphics[width=\linewidth]{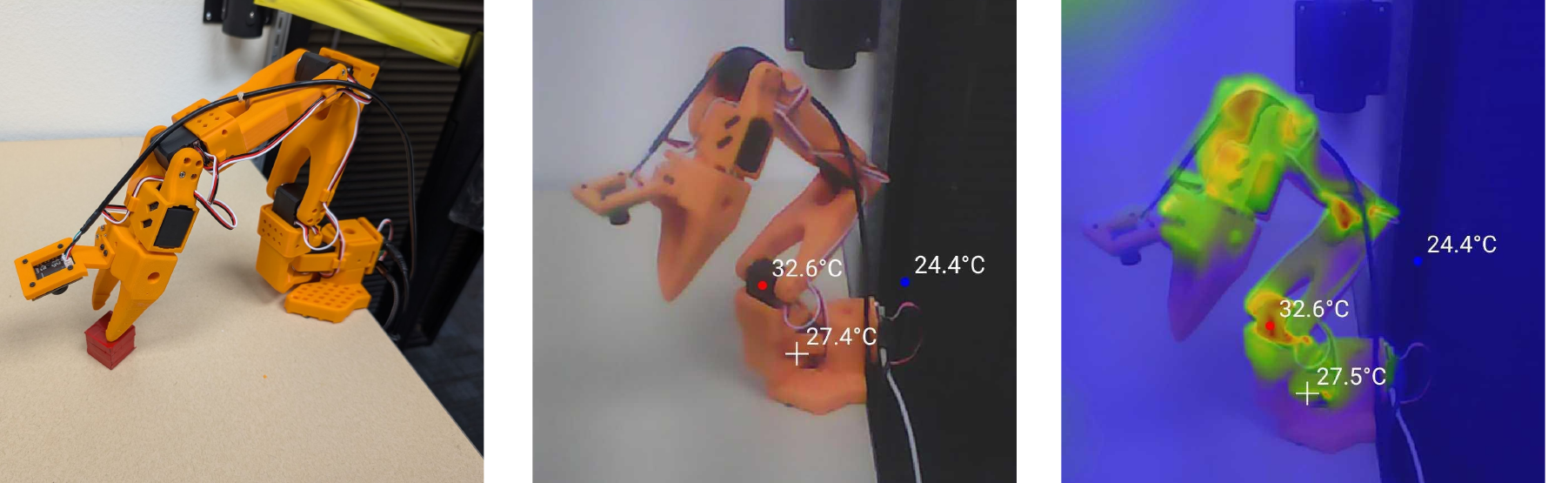}
\caption{\textbf{Physical SO-101 platform.} (a) The arm used for cube lifting. (b) Thermal-camera spot readings. (c) The thermal image illustrates spatially nonuniform heating. The photographed readings illustrate the setup, rather than the heated evaluation thresholds.}
\label{fig:real_so101}
\end{figure}

The physical evaluation in Section~\ref{sec:rq5} uses the SO-ARM101 five-DoF arm for cube lifting. The behavior-cloned base policy and TeAR are trained in simulation and deployed without on-robot fine-tuning. Heating is induced by operating the arm under load. The three evaluated conditions are cool operation below $45^\circ$C, shoulder heating above $50^\circ$C, and heating of both shoulder and wrist above $50^\circ$C, with $20$ trials per method and condition.

TeAR preserves all $20/20$ successful lifts under cool operation and improves success under both heating conditions: from $15/20$ to $17/20$ with shoulder heating, and from $14/20$ to $17/20$ with shoulder-and-wrist heating. These results demonstrate transfer of the simulation-trained adapter to thermal stress on the physical arm, without on-robot fine-tuning. Current and voltage stress are evaluated separately in simulation. The hardware cool-condition label is an experimental temperature range, whereas exact action pass-through depends on the measured telemetry satisfying the gate thresholds in Appendix~\ref{app:hp}.

\subsection{Severity-tier curriculum ablation}
\label{app:severity}

This supplementary study uses BC-Transformer--Can, BC--Threading, and BC--Stack, with $20$ episodes per condition and evaluation seed $42$. Each adapter is trained with the first $n$ tiers of Table~\ref{tab:app_tiers}. We report the mean success rate over hot, stall, brownout, $\mathcal T_{\mathrm{mod}}$, $\mathcal{TC}_{\mathrm{mod}}$, $\mathcal{TV}_{\mathrm{mod}}$, and $\mathcal{TCV}_{\mathrm{mod}}$, then average equally over the three policy--task combinations. These supplementary means are separate from the three-seed comparisons in the main paper.

\begin{table}[t]
\centering\footnotesize
\setlength{\tabcolsep}{4pt}
\caption{\textbf{Severity coverage during training.} Stressed success averaged over three cells and seven conditions. Differences are percentage points relative to the four-tier recipe.}
\label{tab:supp_severity}
\begin{tabular}{lrr}
\toprule
SFT curriculum & Success (\%) & $\Delta$ vs.\ four tiers\\
\midrule
Clean only & $23.8$ & $-13.1$\\
Clean + mild & $39.5$ & $+2.6$\\
Clean + mild + moderate & $40.7$ & $+3.8$\\
Four tiers (reference) & $36.9$ & $0.0$\\
\bottomrule
\end{tabular}
\end{table}

All three curricula that include stressed telemetry outperform clean-only training, improving the observed mean by $13.1$--$16.9$ percentage points. The four-tier reference covers the full training range, including severe telemetry; the two- and three-tier variants have slightly higher observed means on these cells. The consistent gain over clean-only training supports including stressed telemetry in supervision.

\subsection{Alternative adapter trunks}
\label{app:archs}

The supplementary architecture study covers six policy--task combinations: BC on Lift, Can, Threading, and Stack, and BC-Transformer on Can and Square. Each saved configuration uses $20$ episodes per condition and seed $42$. Table~\ref{tab:supp_adapter_archs} averages the seven conditions in Appendix~\ref{app:severity}; the standard deviation is across cells. The reference Transformer achieves the highest mean in this comparison across six policy--task combinations, exceeding the alternatives by $2.7$--$9.0$ percentage points. The main paper reports a separate three-seed architecture study using different trained configurations.

\begin{table}[t]
\centering\footnotesize
\caption{\textbf{Alternative adapter architectures.} Mean success and sample standard deviation across six cells; changes are percentage points from the reference.}
\label{tab:supp_adapter_archs}
\begin{tabular}{lrr}\toprule
Trunk & Success (\%) & $\Delta$\\\midrule
Bottleneck & $26.8\pm21.7$ & $-8.2$\\
Cross-attention & $32.3\pm24.4$ & $-2.7$\\
FiLM & $28.2\pm22.0$ & $-6.8$\\
Hybrid & $26.0\pm20.9$ & $-9.0$\\
Per-index MLP & $29.3\pm19.3$ & $-5.7$\\
Global MLP & $28.8\pm21.1$ & $-6.2$\\
Mixture of experts & $27.5\pm19.8$ & $-7.5$\\
Reference Transformer & $35.0\pm23.1$ & $+0.0$\\
\bottomrule\end{tabular}
\end{table}

\subsection{Hyperparameter sensitivity grid}
\label{app:hp_grid}

\begin{figure}[t]
\centering
\includegraphics[width=\linewidth]{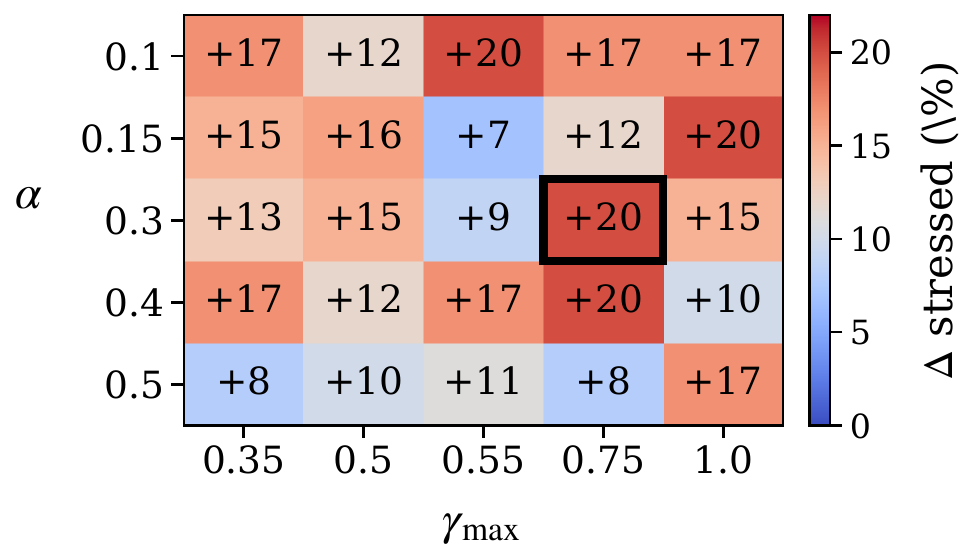}
\caption{\textbf{Selection of correction bounds.} Mean stressed success-rate gain over the frozen base (percentage points), reproduced from the main paper. The outlined cell marks the selected configuration, $(\alpha,\gamma_{\max})=(0.30,0.75)$.}
\label{fig:app_hp_grid}
\end{figure}

We select the reference setting $\alpha=0.30$ and $\gamma_{\max}=0.75$ from the $5\times5$ heatmap in Fig.~\ref{fig:app_hp_grid}. The sweep evaluates the residual and gain bounds on BC-Transformer--Square, using $20$ episodes per condition and seed $42$. Each entry reports the mean success-rate gain over the frozen base across the seven stressed conditions in Appendix~\ref{app:severity}, using the same heatmap as the main paper. The outlined cell marks the selected setting.

\subsection{Inference-time telemetry-channel masking}
\label{app:channel_mask}

We evaluate the full trained adapter with selected channels replaced by nominal values: $T=30^\circ$C, $C=0.30$, or $V=0.96$. Masking changes the telemetry supplied to both the learned adapter and its gate; it does not change the environment's degradation. No adapter is retrained for this study.

The ten policy--task combinations are BC on Can, Square, Stack, and Threading; BCQ on Square and Stack; BC-Transformer on Square; HBC on Square and Threading; and IRIS on Square. Each condition uses $20$ episodes with seed $42$. Table~\ref{tab:supp_channel_mask} averages the same seven stressed conditions listed in Appendix~\ref{app:severity}. Standard deviations describe how the change from full telemetry varies across policy--task combinations.

\begin{table}[t]
\centering\footnotesize
\setlength{\tabcolsep}{5pt}
\caption{\textbf{Telemetry available at inference.} Mean stressed success over ten cells. $\Delta$ is the paired change from full telemetry, reported as mean $\pm$ sample standard deviation across cells, in percentage points.}
\label{tab:supp_channel_mask}
\begin{tabular}{lrr}
\toprule
Visible channels & Success (\%) & $\Delta$ vs.\ full\\
\midrule
T only & $28.3$ & $-4.0\pm4.3$\\
C only & $16.5$ & $-15.8\pm6.7$\\
V only & $19.1$ & $-13.2\pm7.8$\\
T + C & $31.1$ & $-1.1\pm5.3$\\
T + V & $28.8$ & $-3.5\pm4.4$\\
C + V & $18.6$ & $-13.7\pm9.6$\\
\textbf{T + C + V} & $\mathbf{32.3}$ & $0.0$\\
\bottomrule
\end{tabular}
\end{table}

Full telemetry achieves the highest observed mean, with temperature contributing the most among the individual channels: retaining temperature alone loses $4.0$ points, while masking temperature and retaining current and voltage loses $13.7$ points. These results support using all three telemetry channels in TeAR.

\subsection{Per-action correction magnitude analysis}
\label{app:supplementary_corrections}

\begin{figure}[t]
\centering
\includegraphics[width=\linewidth]{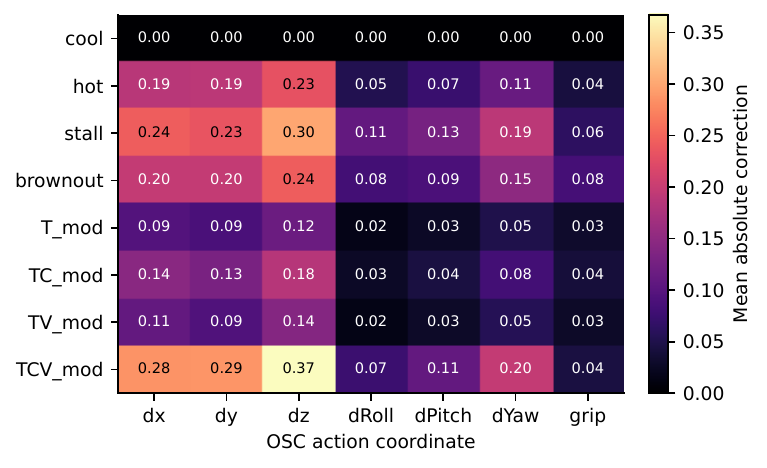}
\caption{\textbf{Mean absolute correction by action coordinate.} Values are recomputed from all ten saved cell summaries. The nominal row is exactly zero; combined stress produces larger corrections, especially in translation. The gripper may receive a learned correction even though the synthetic degradation leaves its command unchanged.}
\label{fig:corrections}
\end{figure}

We measure the mean absolute correction for each of the seven OSC action coordinates using the earlier ten-cell rollout logs. The policy--task combinations match Appendix~\ref{app:channel_mask}; each condition uses $20$ episodes and seed $42$. For each coordinate, we pool rollout steps within each condition and then average equally across cells, as shown in Fig.~\ref{fig:corrections}. The nominal correction is exactly zero. Under stress, corrections are largest on the translation coordinates, particularly the vertical coordinate, and smaller on rotation and the gripper. The measurements describe changes to motion commands, rather than individual joint torques.

\subsection{Curve-family generalization: paired model mismatch}
\label{app:curvefam}

\begin{figure}[t]
\centering
\includegraphics[width=\linewidth]{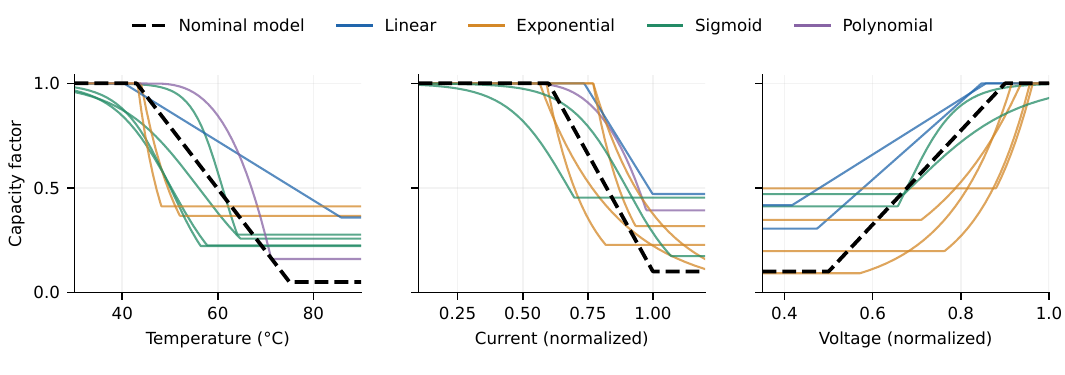}
\caption{\textbf{Evaluated capacity curves.} Colored lines show the eight test curve sets; dashed black lines show the nominal training model. Each set changes all three channels jointly, varying curve shape, onset, and minimum capacity.}
\label{fig:app_curvefam}
\end{figure}

\textbf{Evaluation setup.} We test transfer from the nominal linear training model to eight sets of temperature, current, and voltage curves drawn from linear, exponential, sigmoid, and polynomial families (Fig.~\ref{fig:app_curvefam}; curve seed $20260909$). We evaluate BC--Lift, BC--Can, BC-Transformer--Square, and IRIS--Square under hot, $\mathcal T_{\mathrm{mod}}$, $\mathcal{TC}_{\mathrm{mod}}$, and $\mathcal{TCV}_{\mathrm{mod}}$ conditions. Five episodes per condition with seeds $101$, $202$, and $303$ give $480$ stressed episodes per policy--task combination and method, or $1{,}920$ in total. Horizons are $250$ steps for Lift and $400$ for Can and Square; telemetry is fixed within each episode.

\textbf{Matched comparisons.} Methods share initial conditions, telemetry, and random seeds for policy sampling and disturbances. Under nominal telemetry, $(T,C,V)=(30,0.30,0.96)$, all $240$ correction-method/base comparisons reproduce the base trajectory exactly, verifying nominal pass-through throughout the rollout.

\textbf{Model information.} TeAR learns from the nominal capacity model during training, while the assumed-model inverse applies that model directly at deployment. Neither receives the actual test capacity curves. The privileged inverse receives those curves but does not cancel disturbances. DR-TeAR trains with $64$ randomized curve sets and tests on new parameters and combinations from represented families. Can variants also differ in action-input noise and Lift variants in base-policy training settings, so this comparison evaluates the trained variants rather than curve randomization alone.

\end{document}